\documentclass[journal,comsoc]{IEEEtran}
\usepackage[T1]{fontenc}

\ifCLASSINFOpdf
\else
\fi
\usepackage{amsmath}
\usepackage[cmintegrals]{newtxmath}

\usepackage{times}
\usepackage{soul}
\usepackage{url}
\usepackage[hidelinks]{hyperref}
\usepackage[utf8]{inputenc}
\usepackage{graphicx}
\usepackage{subcaption}
\usepackage{amsmath}
\usepackage{booktabs}
\usepackage{algorithm}
\usepackage{algorithmic}
\usepackage{pifont}
\usepackage{bbm}
\usepackage{cite}
\usepackage{color}
\usepackage{multirow}
\usepackage{multicol}

\newtheorem{proof}{Proof}

\begin{document}
%
\title{Riemannian Self-Attention Mechanism \\ for SPD Networks}
%
%
%

\author{Rui Wang,
        Xiao-Jun Wu*,
        Hui Li,
        and~Josef Kittler,~\IEEEmembership{Life~Member,~IEEE}
\thanks{R. Wang, X.-J. Wu (\textit{Corresponding author}), and H. Li are with the School of Artificial Intelligence and Computer Science, Jiangnan University, Wuxi 214122, China. R. Wang, X.-J. Wu, and H. Li are also with Jiangsu Provincial Engineering Laboratory of Pattern Recognition and Computational Intelligence, Jiangnan University e-mail: \{cs\_wr, wu\_xiaojun, huili.cv\}@jiangnan.edu.cn.}
\thanks{J. Kittler is with the Centre for Vision, Speech and Signal Processing,
University of Surrey, Guildford GU2 7XH, U.K. J. Kittler is also with the School of Artificial Intelligence and Computer Science, Jiangnan University, Wuxi 214122, China. e-mail: j.kittler@surrey.ac.uk.}
}

\markboth{Journal of \LaTeX\ Class Files.}%
{Shell \MakeLowercase{\textit{et al.}}: Bare Demo of IEEEtran.cls for IEEE Communications Society Journals}

\maketitle

\begin{abstract}
Symmetric positive definite (SPD) matrix has been demonstrated to be an effective feature descriptor in many scientific areas, 
as it can encode spatiotemporal statistics of the data adequately on a curved Riemannian manifold, \textit{i.e.}, SPD manifold. 
Although there are many different ways to design network architectures for SPD matrix nonlinear learning, very few solutions explicitly mine the geometrical dependencies of features at different layers. Motivated by the great success of self-attention mechanism in capturing long-range relationships, an SPD manifold self-attention mechanism (SMSA) is proposed in this paper using some manifold-valued geometric operations, mainly the Riemannian metric, Riemannian mean, and Riemannian optimization. Then, an SMSA-based geometric learning module (SMSA-GLM) is designed for the sake of improving the discrimination of the generated deep structured representations. Extensive experimental results achieved on three benchmarking datasets show that our modification against the baseline network further alleviates the information degradation problem and leads to improved accuracy.
\end{abstract}


\IEEEpeerreviewmaketitle

\section{Introduction}

\IEEEPARstart{C}{ovariance} matrices are a well-established tool in many statistical-related fields, yet their usage as data descriptor in the computer vision and pattern recognition (CV\&PR) community is less prevalent. Nonetheless, their efficacy has been proven in diverse applications. In the realm of medical imaging, covariance matrices are used for the classification of time-series data for Brain-Computer Interfaces (BCI) \cite{bci} and for the analysis of magnetic resonance imaging (MRI) \cite{dai,manifoldnet}. In visual classification, the ability of a global covariance matrix to capture the spatiotemporal variations of data points with different lengths has made covariance features highly effective in a number of practical scenarios, including dynamic scene classification \cite{dmk,symnet,sdsma}, facial emotion recognition \cite{spdnet,spdnetbn,spdnetrp}, face recognition \cite{leml,herml,spdml}, and action recognition \cite{jdrml,hgrnet,dmtnet}, \textit{etc}.

However, processing and classifying SPD matrices is challenging, primarily because their underlying space is a curved Riemannian manifold known as the SPD manifold \cite{ArsV}. As a result, conventional Euclidean methods cannot be directly applied. However, Riemannian metrics such as Log-Euclidean Metric (LEM) \cite{ArsV} and Affine-Invariant Riemannian Metric (AIRM) \cite{PenX} have facilitated the extension of Euclidean computations to SPD manifolds. These techniques involve either mapping the manifold to an associated flat space via tangent approximation \cite{pde,mcc,stcd} or embedding it into a Reproducing Kernel Hilbert Space (RKHS) using Riemannian kernel functions \cite{cdl,vem,rsr,rcdl}. Compared with Euclidean metrics, Riemannian metrics are less affected by outliers and noise \cite{eeg-bci}. Additionally, metrics defined on Riemannian manifolds possess various invariance properties, such as affine-invariance or similarity-invariance \cite{ArsV}, which strengthen the generalization ability of learning models to complex data. Nevertheless, the approaches introduced above may yield unsatisfactory results as their feature transforming process could potentially alter the intrinsic geometrical structure of the input data manifold. To address this issue, geometry-aware discriminant analysis algorithms \cite{leml,cml,spdml,ps3d} have been developed to learn a more effective manifold-to-manifold embedding mapping that fully preserves the original Riemannian geometry. Unfortunately, although this type of approach has demonstrated comparatively good performance, the shallow linear mapping function defined on the nonlinear manifolds will inevitably prevent these methods from mining fine-grained representations. 

Convolutional neural networks (ConvNets) \cite{resnet,vgg} are widely recognized for their superiority over conventional shallow learning architectures in terms of learning powerful features. This is not only due to their capacity to perform multi-stage nonlinear computations but also stem from the effectiveness and scalability of gradient-descent training procedure used in backpropagation. Based on this observation, certain researchers have endeavored to extend the paradigm of ConvNets to the context of Riemannian manifolds, bringing new dynamism to visual modeling and learning. SPDNet \cite{spdnet} is an end-to-end Riemannian neural network that introduces a deep and nonlinear learning mechanism for SPD matrices. It consists of several trainable blocks, each comprising an SPD matrix bilinear mapping layer (akin to a convolutional layer) and an eigenvalue regularization layer (similar to ReLU) for data compression and nonlinear activation. This innovative architecture ensures that the SPD properties of the input data are maintained layer-by-layer. Subsequently, a Riemannian-Euclidean mapping layer is employed to embed the obtained Riemannian representations into an Euclidean space to enable classification. This design has served as the foundation for further advancements \cite{spdnetbn,dmtnet,manifoldnet,symnet} in refining and customizing the existing building blocks to better accommodate diverse CV\&PR tasks. Theoretically speaking, Riemannian deep learning has the potential to extract more effective geometric features for enhanced visual classification compared to the aforementioned SPD matrix learning methods.

The utilization of deep learning techniques in the field of SPD manifolds is promising, but it remains in its infancy. SPD neural networks encounter a major limitation in which some critical structural information will be lost during multi-stage data compression mapping, which hinders their ability to learn informative deeper representations. Recent research \cite{vgg,resnet} has shown that the depth of a neural network is crucial for achieving good performance. However, when it comes to SPD networks, the aforementioned degradation problem renders it impossible to improve the classification accuracy by simply increasing the number of operation layers. This is also verified in Fig. 1 of \cite{sdsma}. However, to the best of our knowledge, there is limited research available regarding the issue of information degradation of SPD networks. A typical solution to this problem is presented in \cite{sdsma}. To be specific, on the tail of the backbone network (SPDNet), the authors establish a stacked SPD manifold autoencoder (SSMAE) to enrich the depth of structured features. By utilizing the reconstruction error terms associated with SSMAE and each SPD manifold autoencoder (SMAE), the network embedding function can asymptotically approximate an identity mapping, not only improving the classification performance but also simplifying the training of deeper network. Even so, as the number of layers continues to deepen, the classification accuracy will slowly decline after reaching a peak. The essential reason is that with the increase of SMAE transformation stages, a series of SPD matrix upsampling (resulting in rank-deficient matrices) and activation operations will unavoidably amplify the distortion of raw statistics.

\begin{figure*}[!t]
 \centering
 \includegraphics[scale=0.58]{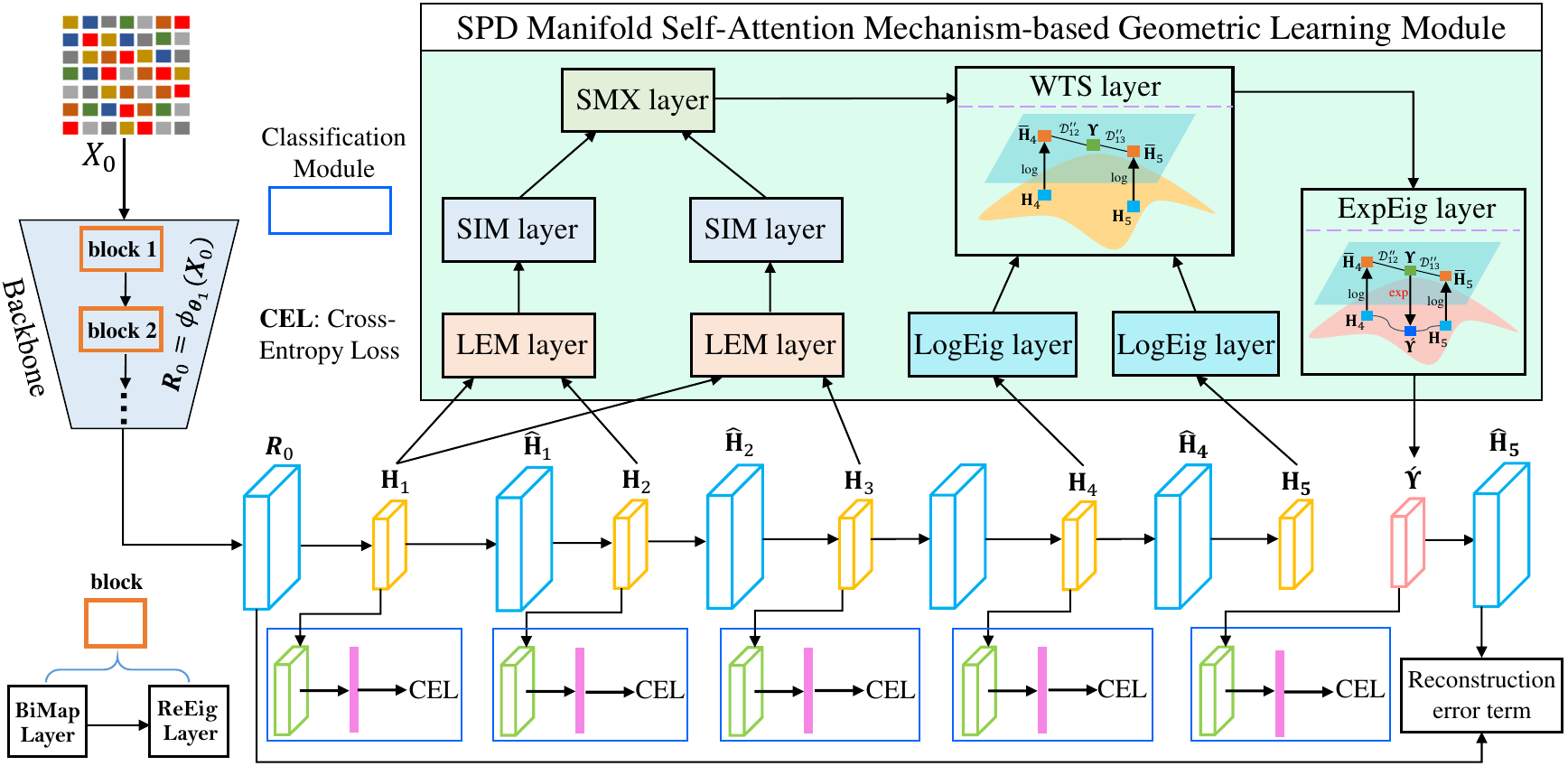}
 \caption{The schematic diagram of the designed SPD manifold self-attention mechanism-based geometric learning module (SMSA-GLM) and its combination with DSPDNet \cite{sdsma}. In this figure, we set the number $\rm{E}$ of the stacked SMAEs to $5$ as an example. The SMSA-GLM makes up of the layers of LEM, SIM, SMX, LogEig, WTS, and ExpEig for the implementation of similarity measurement and weighted average in the context of SPD manifolds, which will be introduced in detail later. In this article, we use $\boldsymbol{\rm{H}}_i$ and $\hat{\boldsymbol{\rm{H}}}_i$ to denote the data generated by the hidden layer and upsampling layer of the $i^{\rm{th}}$ ($i=1\to \rm{E}$) SMAE, respectively. Besides, $\boldsymbol{R}_0$ and $\acute{\boldsymbol{\Upsilon}}$ respectively represent the input and output data of SSMAE and SMSA-GLM.}
 \label{fig-1}
\end{figure*}

Given the great success of self-attention mechanism defined in the Euclidean space in characterizing the long-range relationships between features \cite{vit,ovit,gsma}, we propose a self-attention mechanism on the SPD manifolds (SMSA) to address the aforementioned issue by referring to some geometric operators, including Riemannian metric, Riemannian mean, and Riemannian optimization. Considering that the information flow between different SMAEs of the deeper SPD network (for simplicity, we call it DSPDNet) \cite{sdsma} is relatively simple, a geometric learning module based on the proposed SMSA (SMSA-GLM) is designed for DSPDNet with the purpose of mining the statistical complementarity inherent in 
the multiple geometric features to guide the generation of more discriminative deep representations. However, in contrast to regular self-attention that operates on vectors, our approach regards SPD matrices that reside on an SPD manifold as query, key, and value. In such a case, in order to achieve effective mining and aggregation of geometrical dependencies between different depth features, we first exploit the Riemannian metric instead of the commonly used dot-product to measure the similarity between query and key. On the basis of the attention matrix computed, the weighted Fréchet Mean (weighted Riemannian mean) is naturally utilized to obtain the final outputs. There are three main reasons for using Riemannian mean: 1) it has shown significant theoretical and practical advantages in Riemannian data analysis \cite{PenX}; 2) it is faithful to the Riemannian geometry of SPD manifolds; 3) it introduces very few parameters to be learned. Fig. \ref{fig-1} is an overview of the proposed method. 

The advantages of the proposed method are exemplified on three visual classification tasks, namely facial emotion recognition, skeleton-based hand action recognition, and skeleton-based action recognition with unmanned aerial vehicles (UAVs). The experimental results achieved on three benchmarking datasets show that the suggested SMSA-GLM can help the baseline network to further overcome the problem of information degradation, producing higher accuracy than the previous methods.

\section{Related Works}
The preceding statement acknowledges the noteworthy advancements that conventional geometry-aware Riemannian learning techniques have achieved in SPD matrix-based visual processing and classification. However, their inherent shallow linear embedding mechanism proves inadequate for extracting fine-grained geometric features, particularly in complex visual scenarios. Taking inspiration from the benefits of ConvNets, Huang et al. \cite{spdnet} create an end-to-end Riemannian neural network that specializes in non-linear learning of SPD matrices. This network is distinct from ConvNets in that it processes structured matrices, with SPD matrices as input and the output matrices are also SPD. 
By extending the philosophy of SPDNet \cite{spdnet} to the context of Gaussian embedded Riemannian manifolds, Nguyen et al. \cite{hgrnet} design two Gaussian aggregation sub-networks for both spatial and temporal domains. This results in the generation of more informative SPD matrices w.r.t the input skeletal data for action recognition. Brooks et al. \cite{spdnetbn} incorporate a Riemannian batch normalization layer, implemented by parallel transport and Riemannian barycenter, into each building block of SPDNet to facilitate weight optimization and network training. To better adapt to visual tasks based on small-scale datasets, Wang et al. \cite{symnet} design a lightweight cascaded Riemannian network for SPD matrix encoding and classification. This network exploits a shallow optimization algorithm for parameter learning, enabling it to achieve comparable accuracy and superior computational efficiency than other competing methods. 

Nevertheless, the Riemannian deep learning approach introduced above is a generalization of the Euclidean fully connected network in the domain of Riemannian manifolds, which fails to mine the local geometrical structure of the manifold-valued data. In light of the crucial role that local data structures play in producing impactful representations, Zhang et al. \cite{dmtnet} design a 2D convolutional layer to compress SPD features. Notably, this approach mandates that each convolutional kernel itself must also be SPD. Different from the operating mechanism proposed in \cite{dmtnet}, Chakraborty et al. \cite{manifoldnet} utilize the weighted Fréchet Mean to implement the weighted sums of Euclidean convolution on the grids of manifold-valued data like diffusion tensors (each pixel in the data grid represents a Riemannian matrix). It can be noted that this method ignores the effective learning of the statistics embodied in each pixel. To fully abide by the principle of Riemannian geometry, Chen et al. \cite{msnet} define the local information on the SPD manifolds from the perspective of category theory, and design a submanifold network for mining local geometry.  

\section{Proposed Method}
Plenty of experimental evidences suggest that carrying out deep learning on the SPD manifolds can produce more effective geometric features for improved classification compared to Riemannian shallow learning methods. However, achieving accuracy gains is not as straightforward as increasing the number of operation layers, owing to the loss of main structural information during multi-stage data compressed sensing. As far as we know, there are few studies on the degradation problem of SPD networks in the research community, among which the method proposed by Wang et al. \cite{sdsma} is a more effective one that pays attention to this issue. The primary ideology behind the designed DSPDNet is to ensure that the embedding function of the model established at the end of the shallower backbone closely resembles an identity mapping. In fact, this does not mean that the depth of DSPDNet can be deepened infinitely. Experimental results show that there is a performance upper limit. In this article, we design an SMSA-GLM for DSPDNet to improve its learning capacity. 


\subsection{Preliminaries}

\textbf{SPD Manifold and the Corresponding Log-Euclidean Metric:} For all non-zero vector $\boldsymbol{v}\in\mathbbm{R}^d$, a real-valued matrix $\boldsymbol{\mathrm{X}}$ is called SPD if and only if $\boldsymbol{v}^{\rm{T}}\boldsymbol{\mathrm{X}v}>0$. The space spanned by a family of $d$-by-$d$ SPD matrices is a commutative Lie group, with a manifold structure denoted as $\boldsymbol{\mathcal{S}}_{++}^d$. More formally:
\begin{equation}
\begin{split}
    \boldsymbol{\mathcal{S}}_{++}^d := \{\boldsymbol{\mathrm{X}}\in\mathbbm{R}^{d\times d}:\,&\boldsymbol{\mathrm{X}}=\boldsymbol{\mathrm{X}}^{\rm{T}},\,\\
    &\boldsymbol{v}^{\rm{T}}\boldsymbol{\mathrm{X}v}>0,\forall \boldsymbol{v}\in\mathbbm{R}^d\backslash\{0_d\}\}.
\end{split}
\end{equation}
Therefore, the concepts of differential geometry, such as geodesic, can be utilized to address 
$\boldsymbol{\mathcal{S}}_{++}^d$. Moreover, any bi-invariant metric $\langle \,,\rangle$ on the Lie group of SPD matrices corresponds to an Euclidean metric in the SPD matrix logarithmic domain, \textit{i.e.}, the tangent space at identity matrix $\boldsymbol{\mathcal{T}_I\mathcal{S}}_{++}^d$, it is also called the Log-Euclidean Metric. 

Specifically, for any two tangent elements $\boldsymbol{T}_i,\boldsymbol{T}_j$, their scalar product in $\boldsymbol{\mathcal{T}}_{\boldsymbol{\mathrm{X}}}\boldsymbol{\mathcal{S}}_{++}^d$ is given as:
\begin{equation}
\langle \boldsymbol{T}_i,\boldsymbol{T}_j \rangle_{\boldsymbol{\mathrm{X}}} = \langle D_{\boldsymbol{\mathrm{X}}}\mathrm{log}_{\cdot}\boldsymbol{T}_i, D_{\boldsymbol{\mathrm{X}}}\mathrm{log}_{\cdot}\boldsymbol{T}_j \rangle_{\boldsymbol{I}},
\label{eq1}
\end{equation}
where $D_{\boldsymbol{\mathrm{X}}}\mathrm{log}_{\cdot}\boldsymbol{T}$ signifies the directional derivative of the matrix logarithm at $\boldsymbol{\mathrm{X}}$ along $\boldsymbol{T}$. 
The logarithmic mapping with respect to the Riemannian metric is defined by the matrix logarithms:
\begin{equation}
\mathrm{log}_{\boldsymbol{\mathrm{X}}_i}(\boldsymbol{\mathrm{X}}_j) = D_{\mathrm{log}(\boldsymbol{\mathrm{X}}_i)}\mathrm{exp}_{\cdot}(\mathrm{log}(\boldsymbol{\mathrm{X}}_j) - \mathrm{log}(\boldsymbol{\mathrm{X}}_i)).
\label{eq2}
\end{equation}
On the basis of the differentiation of $\mathrm{log}\circ\mathrm{exp}=\boldsymbol{I}$, we can obtain $D_{\mathrm{log}_{\boldsymbol{\mathrm{X}}}}\mathrm{exp}_{\cdot} = (D_{\boldsymbol{\mathrm{X}}}\mathrm{log}_{\cdot})^{-1}$. Similarly, the matrix exponential mapping can be expressed as:
\begin{equation}
\mathrm{exp}_{\boldsymbol{\mathrm{X}}_i}(\boldsymbol{T}_j) = \mathrm{exp}(\mathrm{log}(\boldsymbol{\mathrm{X}}_i) + D_{\boldsymbol{\mathrm{X}}_i}\mathrm{log}_{\cdot}\boldsymbol{T}_j).
\label{eq3}
\end{equation}
Combining Eq. (\ref{eq1}), Eq. (\ref{eq2}), and Eq. (\ref{eq3}), the LEM can be formulated as:   
\begin{equation}
\begin{split}
D_{LEM}(\boldsymbol{\mathrm{X}}_i,\boldsymbol{\mathrm{X}}_j) &= \langle {\rm log}_{\boldsymbol{\mathrm{X}}_i}(\boldsymbol{\mathrm{X}}_j),{\rm log}_{\boldsymbol{\mathrm{X}}_i}(\boldsymbol{\mathrm{X}}_j) \rangle_{\boldsymbol{\mathrm{X}}_i} \\
& = || {\rm log}(\boldsymbol{\mathrm{X}}_j) - {\rm log}(\boldsymbol{\mathrm{X}}_i) ||_{\rm{F}}.
\end{split}
\label{lem}
\end{equation}
It can be found that compared with AIRM, LEM works directly in the domain of SPD matrix logarithms, demonstrating comparatively high computational efficiency. As a consequence, we choose it as the distance measure in this paper. 
For detailed information, please kindly refer to \cite{ArsV}. 

\textbf{Data Modeling with Second-Order Statistics:} Let $\boldsymbol{S}_i=[\boldsymbol{s}_1,\boldsymbol{s}_2,...,\boldsymbol{s}_{n_i}]$ denotes a given data sequence (image set or video clip) containing $n_i$ entries, where $\boldsymbol{s}_r \in \mathbbm{R}^{d\times1}$ represents the $r^{\rm{th}}$ vectorized instance. The second-order representation corresponding to $\boldsymbol{S}_i$ can be computed as follows:
\vskip -0.05in
\begin{equation}
   \boldsymbol{\mathrm{X}}_i=\frac{1}{n_i-1}\sum_{r=1}^{n_i}(\boldsymbol{s}_r-\boldsymbol{u}_i)(\boldsymbol{s}_r-\boldsymbol{u}_i)^{\rm{T}}.
\label{e1}
\end{equation}
where $\boldsymbol{u}_i$ is the mean of $\boldsymbol{S}_i$, computed by $\boldsymbol{u}_i=\frac{1}{n_i}\sum_{r=1}^{n_i}\boldsymbol{s}_r$. Given that $\boldsymbol{\mathrm{X}}_i$ is not necessarily positive definite, the following common regularization strategy is imposed on it: $\boldsymbol{\mathrm{X}}_i \gets  \boldsymbol{\mathrm{X}}_i+\lambda\boldsymbol{I}_d$, where $\boldsymbol{I}_d$ is a  $d$-by-$d$ identity matrix, and $\lambda$ is set to $trace( \boldsymbol{\mathrm{X}}_i)\times 10^{-3}$ in all the experiments. At this time, $\boldsymbol{\mathrm{X}}_i$ has become a real SPD manifold-valued element \cite{PenX}.

\subsection{The Basic Layers of SPD Neural Network}
Let $\boldsymbol{\mathrm{X}}_{k-1}\in\boldsymbol{\mathcal{S}}_{++}^{d_{k-1}}$ be the input SPD matrix of the $k^{\rm{th}}$ layer, the introduction to the Riemannian operation layers for processing the SPD matrix is as follows: 

\textbf{BiMap Layer}: This layer can be thought of as a variation of the standard dense layer, where the input SPD matrices are compressed into lower-dimensional ones using a bilinear mapping function $f_b$, expressed as: $\boldsymbol{\mathrm{X}}_k=f_b^{(k)}(\boldsymbol{W}_k,\boldsymbol{\mathrm{X}}_{k-1})=\boldsymbol{W}_k^{\rm{T}}\boldsymbol{\mathrm{X}}_{k-1}\boldsymbol{W}_k$, where $\boldsymbol{W}_k\in\mathbbm{R}^{d_{k-1}\times d_k}$ ($d_k<d_{k-1}$) is the transformation matrix to be learned. To ensure that $\boldsymbol{\mathrm{X}}_k$ lies in another SPD manifold $\boldsymbol{\mathcal{S}}_{++}^{d_k}$, $\boldsymbol{W}_k$ needs to be column full-rank. In addition, it is necessary to impose semi-orthogonality constraint on $\boldsymbol{W}_k$, which results in a compact Stiefel manifold $St(d_k, d_{k-1})$ for the weight space \cite{ArsV}. By optimizing $\boldsymbol{W}_k$ over this manifold, it becomes possible to yield optimal solutions.

\textbf{ReEig Layer}: This layer bears resemblance to the conventional ReLU layers, and its purpose is to introduce nonlinearity for DSPDNet by using a rectification function $f_r$ to regularize the small positive eigenvalues of each input SPD matrix. Its definition is given below: $\boldsymbol{\mathrm{X}}_k=f_r^{(k)}(\boldsymbol{\mathrm{X}}_{k-1})=\boldsymbol{U}_{k-1}\mathrm{max}(\epsilon\boldsymbol{I},\boldsymbol{\Sigma}_{k-1})\boldsymbol{U}_{k-1}^{\rm{T}}$, where $\epsilon$ is a small activation threshold, and $\boldsymbol{\mathrm{X}}_{k-1}=\boldsymbol{U}_{k-1}\boldsymbol{\Sigma}_{k-1}\boldsymbol{U}_{k-1}^{\rm{T}}$ refers to the eigenvalue decomposition. The ReEig operation serves to protect the matrices from degeneration, as is evident.

\textbf{LogEig Layer}: This layer is designed to enable the Euclidean learning methods to be applicable to the generated manifold-valued network features. It is implemented by imposing Riemannian computation on the input SPD matrices using the logarithmic mapping function $f_l$, formulated as: $\boldsymbol{\mathrm{X}}_k=f_l^{(k)}(\boldsymbol{\mathrm{X}}_{k-1})=\boldsymbol{U}_{k-1}\mathrm{log}(\boldsymbol{\Sigma}_{k-1})\boldsymbol{U}_{k-1}^{\rm{T}}$. Here, $\boldsymbol{\mathrm{X}}_{k-1}=\boldsymbol{U}_{k-1}\boldsymbol{\Sigma}_{k-1}\boldsymbol{U}_{k-1}^{\rm{T}}$ refers to the eigenvalue decomposition, and ${\rm{log}}(\boldsymbol{\Sigma})$ denotes the logarithm operation applied to each diagonal element of $\boldsymbol{\Sigma}_{k-1}$.

\subsection{SPD Manifold Self-Attention Mechanism (SMSA)}
\label{smsa}
Let $Q=\{q_1,q_2,...,q_{N_Q}\}$, $K=\{\kappa_1,\kappa_2,...,\kappa_{N_{K}}\}$, and $V=\{v_1,v_2,...,v_{N_V}\}$ respectively denote the query set, key set, and value set composed of a series of SPD matrices with the same dimensionality. To implement SMSA, an essential step is to compute the similarity between $q_i$ and $\kappa_j$. In the Euclidean space, there are several means to measure the similarity, among which dot-product \cite{ain} is widely used. However, unlike standard self-attention that processes vectors, our method regards SPD matrices that lie in an SPD manifold as query, key, and value. Therefore, we make use of LEM defined in Eq. (\ref{lem}) to determine the similarity between SPD manifold-valued query and key. 

For effective weighting operation, the obtained distances can not be directly used as the attention weights. The reason is that the higher the similarity between any two samples, the smaller the corresponding distance. For $q_i$ and $\kappa_j$ ($i$=$1\to n_q$, $j$=$1\to n_{\kappa}$), the similarity between them is defined as:
\begin{equation}
    SIM(q_i,\kappa_j) = \frac{1}{1+log[1+D_{LEM}^2(q_i,\kappa_j)]}.
\end{equation}
It is evident that $SIM(,)$ is a strictly decreasing function w.r.t $D_{LEM}(,)$. At this point, the weight matrix can be expressed as:
\begin{equation}
    \boldsymbol{A} = 
    \begin{bmatrix}
    a_{i1} & \cdots\ & a_{i N_K} \\
    \vdots & \ddots & \vdots \\
    a_{N_{Q}1} & \cdots\ & a_{N_Q N_K} \\
    \end{bmatrix},
\end{equation}
where $a_{ij}:=SIM(q_i,\kappa_j)$. Considering that the values of each row in $\boldsymbol{A}$ is not necessarily convex, the $Softmax$ function is therefore exploited to narrow the value range along the row direction. Then, the new weight matrix is given below:
\begin{equation}
    \tilde{\boldsymbol{A}} =Softmax(\boldsymbol{A})= 
    \begin{bmatrix}
    \tilde{a}_{i1} & \cdots\ & \tilde{a}_{i N_K} \\
    \vdots & \ddots & \vdots \\
    \tilde{a}_{N_Q 1} & \cdots\ & \tilde{a}_{N_Q N_K} \\
    \end{bmatrix}.
\end{equation}

To implement self-attention on the SPD manifolds, another necessary step is to perform weighting operation. Taking the non-Euclidean geometry of $\boldsymbol{v}_i$ ($i=1\to N_V$) into account, we treat the Fréchet Mean (Riemannian mean) as the final output feature matrix. Specifically, given a set of SPD matrices $\{\boldsymbol{\mathrm{X}}_i\}_{i=1}^{N_*}$, its Riemannian mean ($\boldsymbol{P}^*$) can be formulated as \cite{ArsV}:
\begin{equation}    \boldsymbol{P}^*=\mathop\mathrm{arg\;min}\limits_{\boldsymbol{P}\in\boldsymbol{\mathcal{S}}^{d_{*}}_{++}}\sum_{i=1}^{N_*}{D}_{\mathcal{M}}^2(\boldsymbol{\mathrm{X}}_i,\boldsymbol{P}),
\label{rm}
\end{equation}
where $D_{\mathcal{M}}(,)$ signifies any SPD manifold-valued Riemannian metric. However, the solution of Eq. (\ref{rm}) does not have a closed form when the SPD manifold is endowed with AIRM. In this case, Eq. (\ref{rm}) needs to be estimated using an iterative strategy until terminating conditions are reached \cite{manifoldnet}. On the contrary, once the SPD manifold is equipped with LEM, a closed-form solution of Eq. (\ref{rm}) can be derived. 
\;\\

\textbf{Theorem 1.} \textit{The Riemannian mean of a set of SPD matrices $\{\boldsymbol{\mathrm{X}}_i\}_{i=1}^{N_*}$ $\mathrm{w.r.t}$ LEM is computed by:}
\begin{equation}
    \boldsymbol{{P}}^*={\rm{exp}}\left[\frac{1}{n}\sum_{i=1}^{N_*}{\rm{log}}(\boldsymbol{\mathrm{X}}_i)\right].
    \label{lem-rm}
\end{equation}
\begin{proof}
Eq. (\ref{lem-rm}) can be obtained via $\frac{\partial \sum_{i=1}^{N_*}D_{LEM}^2(\boldsymbol{\mathrm{X}}_i,\boldsymbol{P})}{\partial \boldsymbol{P}}$=0.
\end{proof}
\;\\
To realize the weighting operation, we extend the aforementioned definition of Riemannian mean to a weighted form, also known as weighted Fréchet Mean:
\begin{equation}    \boldsymbol{P}^*=\mathop\mathrm{arg\;min}\limits_{\boldsymbol{P}\in\boldsymbol{\mathcal{S}}^{d_{*}}_{++}}\sum_{i=1}^{N_*}w_i{D}_{\mathcal{M}}^2(\boldsymbol{\mathrm{X}}_i,\boldsymbol{P}),
\label{wrm}
\end{equation}
where $w_i$ is the weight assigned to each $\boldsymbol{\mathrm{X}}_i$, satisfying $w_i>0$ for all $i\in \{1,2,...,N_*\}$ and $\sum_i w_i=1$. 
\;\\

\textbf{Theorem 2.} \textit{The weighted Riemannian mean of a set of SPD matrices $\{\boldsymbol{\mathrm{X}}_i\}_{i=1}^{N_*}$ $\mathrm{w.r.t}$ LEM is computed by:}
\begin{equation}
    \boldsymbol{{P}}^*={\rm{exp}}\left[\sum_{i=1}^{N_*}w_i{\rm{log}}(\boldsymbol{\mathrm{X}}_i)\right].
    \label{lem-wrm}
\end{equation}
\begin{proof}
Since $\sum_i w_i=1$, Eq. (\ref{lem-wrm}) can be obtained via $\frac{\partial \sum_{i=1}^{N_*}w_iD_{LEM}^2(\boldsymbol{\mathrm{X}}_i,\boldsymbol{P})}{\partial \boldsymbol{P}}$=0.
\end{proof}
\;\\
Given the weight matrix $\tilde{\boldsymbol{A}}$ and the value set $V$, we can follow Eq. (\ref{lem-wrm}) to define the final output feature matrix as:
\begin{equation}
    \boldsymbol{v}_i^{'}={\rm{exp}}\left[\sum_{j=1}^{N_V}\tilde{a}_{ij}{\rm{log}}(\boldsymbol{v}_j)\right],
    \label{lem-wrm-2}
\end{equation}

\subsection{DSPDNet Embedded with SMSA-based Geometric Learning Module (SMSA-GLM)}
\label{smsa-model}
As mentioned before, SMSA-GLM is designed to assist DSPDNet to further mitigate the degradation problem by making use of different depth features to guide the generation of geometric representations with richer statistics. To achieve this, there are three factors needs to be considered in module design: 1) the rule for generating query, key, and value; 2) the forward propagation; 3) the objective function of the whole network.  

\textbf{The Rule for Generation Query, Key, and Value}: Let $\boldsymbol{\mathcal{X}}=[\boldsymbol{\mathrm{X}}_1,\boldsymbol{\mathrm{X}}_2,...,\boldsymbol{\mathrm{X}}_N]$ be the SPD manifold-valued training set and $\boldsymbol{L}=[l_1,l_2,...,l_N]$ be the corresponding label vector. For the $i^{\rm{th}}$ input SPD matrix $\boldsymbol{\mathrm{X}}_i$, the low-dimensional feature matrix produced by the backbone of DSPDNet is expressed as: $\boldsymbol{R}_i=\phi_{\theta_1}(\boldsymbol{\mathrm{X}}_i)$, where $\phi_{\theta_1}$ denotes the nonlinear embedding function and $\theta_1$ indicates the to-be-learnt parameters of the backbone. At the end of the backbone is the established SSMAE for enriching the depth of representations. Since SSMAE makes up of $\mathrm{E}$ SMAEs (the output data of the $(e-1)^{\rm{th}}$ SMAE is used as the input of the $e^{\rm{th}}$ SMAE), we follow \cite{sdsma} to respectively formulate the output feature matrix of the hidden layer and reconstruction layer of the $e^{\rm{th}}$ ($e=1\to \rm{E}$) SMAE as:
\begin{alignat}{1}
   &{\boldsymbol{\rm{H}}}_e(\boldsymbol{R}_i)= f_{b}(\boldsymbol{W}_{e_1},{\hat{\boldsymbol{\rm{H}}}}_{e-1}(\boldsymbol{R}_i))=\boldsymbol{W}_{e_1}^{\rm{T}}\pi({\boldsymbol{\hat{\rm{H}}}}_{e-1}(\boldsymbol{R}_i))\boldsymbol{W}_{e_1},\\
   &{\boldsymbol{\hat{\rm{H}}}}_e(\boldsymbol{R}_i)= f_{b}(\boldsymbol{W}_{e_2},{{\boldsymbol{\rm{H}}}}_{e}(\boldsymbol{R}_i))=\boldsymbol{W}_{e_2}{\boldsymbol{{\rm{H}}}}_{e}(\boldsymbol{R}_i)\boldsymbol{W}_{e_2}^{\rm{T}},
\end{alignat}
where $f_{b}$ is the bilinear mapping function, $\pi$ means the nonlinear activation (ReEig operation in this paper), and  $\boldsymbol{W}_{e_1}\in\mathbbm{R}^{d_{e-1}\times d_e}$ ($d_e\leq d_{e-1}$) and $\boldsymbol{W}_{e_2}\in\mathbbm{R}^{d_{e-1}\times d_e}$ represent the to-be-learnt transformation matrices of the $e^{\rm{th}}$ SMAE.  In the following, we abbreviate $\boldsymbol{\rm{H}}_e(\boldsymbol{R}_i)$ and $\boldsymbol{\hat{\rm{H}}}_{e}(\boldsymbol{R}_i)$ as $\boldsymbol{\rm{H}}_e$ and $\boldsymbol{\hat{\rm{H}}}_{e}$ for simplicity.

\begin{figure*}[!t]
 \centering
 \includegraphics[scale=0.58]{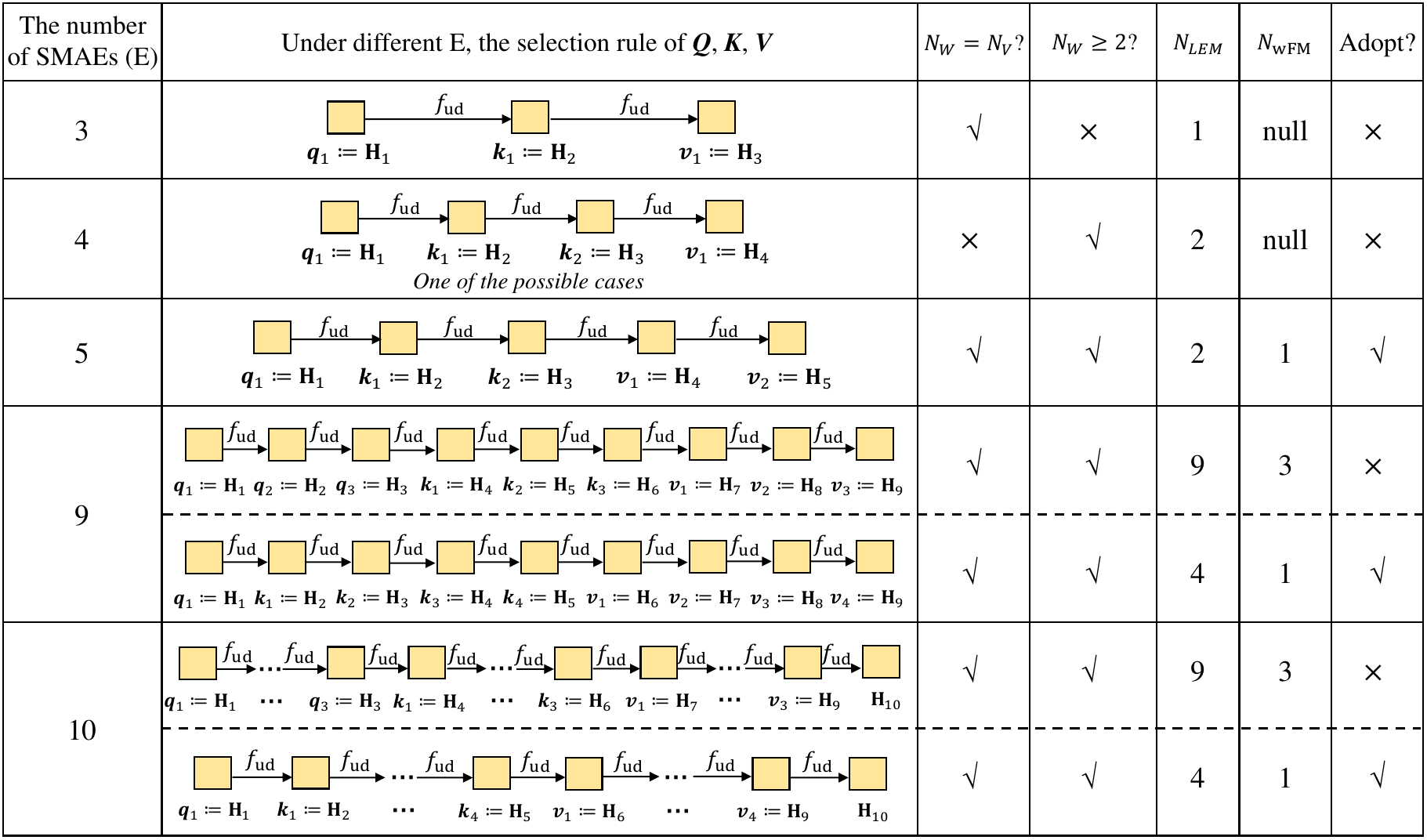}
 \caption{The selection mode of query, key, and value. Wherein, $N_W$ signifies the number of columns of the computed weight matrix, $N_V$ represents the number of values contained in $\boldsymbol{V}$, $N_{LEM}$ and $N_{\rm{wFM}}$ respectively denote the number of LEM and wFM computations involved in SMSA, and $f_{ud}$ represents the transformation function between the hidden layers of any two adjacent SMAEs, consisting of the upsampling, downsampling, and nonlinear activation operations.}
 \label{qkv}
\end{figure*} 

As introduced in Section \ref{smsa}, this article selects LEM for realizing similarity measurement and wFM computation in SMSA. However, in the backpropagation process, LEM involves not only the same eigenvalue operations as the forward pass, but also the computation of inverse matrices. This will lead to a prominent increase in the computational burden of the model as the number of elements in $\boldsymbol{Q}$, $\boldsymbol{K}$, and $\boldsymbol{V}$ increases. As a consequence, we let the constructed SMSA-GLM act directly on all the hidden layers of SSMAE. Fig. \ref{qkv} illustrates the selection mode of $\boldsymbol{Q}$, $\boldsymbol{K}$, and $\boldsymbol{V}$ under different network depth (different $\rm{E}$). From this figure, we can note that when $\rm{E}$ is configured as 3 or 4, the architecture of DSPDNet at this time will not support the implementation of the designed SMSA-GLM. According to Eq. (\ref{lem-wrm-2}), it can be deduced that the necessary condition for the wFM framework to be constructed is that $N_W$ ($N_W \geq 2$) equals to $N_V$. This is the basic reason why the two cases mentioned above can not be adopted. When increasing the value of $\rm{E}$ to 9, we can get two feasible selection rules as listed in Fig. \ref{qkv}. It can be found that $N_{LEM}$ and $N_{\rm{wFM}}$ corresponding to the first selection rule are twice than that of the second one. Of course, the value of $\rm{E}$ can also be an even number. However, the computational burden conveyed by the first line of $\rm{E}=10$ listed in Fig. \ref{qkv} is significantly higher than that of the second line.
Based on the above analysis, the selection protocol of $\boldsymbol{Q}$, $\boldsymbol{K}$, and $\boldsymbol{V}$ is defined as the following three aspects: 1) $N_W=N_V$; 2) the designed SMSA-GLM can maximize the use of all the hidden layers in SSMAE; 3) on the basis of 2), minimizing the number of eigenvalue operations.
As a result, a feasible and time-saving way to combine SMSA-GLM with DSPDNet is to set $\boldsymbol{Q}=\{\boldsymbol{\rm{H}}_1\}$, $\boldsymbol{K}=\{\boldsymbol{\rm{H}}_2,...,\boldsymbol{\rm{H}}_{(\mathrm{E}-x)/2+1}\}$, and $\boldsymbol{V}=\{\boldsymbol{\rm{H}}_{(\mathrm{E}-x)/2+2},...,\boldsymbol{\rm{H}}_{\mathrm{E}-x+1}\}$, respectively. Here, when $\mathrm{E}$ ($\mathrm{E}>4$) is an odd number, $x=1$, otherwise $x=2$.  

\textbf{The Forward Propagation}: To implement the forward pass of the SPD matrix involved in the designed geometric learning module, the following six auxiliary layers are designed by referring to Section \ref{smsa}:

\textit{LEM layer}: This layer is designed to measure the similarity between $\boldsymbol{q}_i$ and $\boldsymbol{k}_j$ via the distance function $f_{LEM}$. According to the previous introduction, we know that $\boldsymbol{Q}=\{\boldsymbol{\rm{H}}_1\}$ and $\boldsymbol{K}=\{\boldsymbol{\rm{H}}_2,\boldsymbol{\rm{H}}_3...,\boldsymbol{\rm{H}}_{h}\}$, where $h=(\mathrm{E}-x)/2+1$. For simplicity, the symbol $h$ is also used in the following. Therefore, based on Eq. (\ref{lem}), this layer can be formulated as:
\begin{equation}
    \mathcal{D}_{1j}=f_{LEM}(\boldsymbol{\rm{H}}_1,\boldsymbol{\mathrm{H}}_j)=||\mathrm{log}(\boldsymbol{\rm{H}}_1)-\mathrm{log}(\boldsymbol{\mathrm{H}}_j)||_{\rm{F}}^2,
    \label{lem-layer}
\end{equation}
where $j=2\to h$.

\textit{SIM layer}: This layer is designed to enable each $\mathcal{D}_{1j}$ computed above to be a valid weight for the subsequent weighting operation via the transformation function $f_{SIM}$, given below:
\begin{equation}
    \mathcal{D}_{1j}^{'} =f_{SIM}(\mathcal{D}_{1j})=\frac{1}{1+log(1+\mathcal{D}_{1j})}.
    \label{sim}
\end{equation}

\textit{SMX layer}: Since the values in $\boldsymbol{\mathfrak{D}}=\{\mathcal{D}_{12}^{'},\mathcal{D}_{13}^{'},...,\mathcal{D}_{1h}^{'}\}$ do not necessarily satisfy convexity, this layer is designed to shrink its value range via the $Softmax$ function $f_{SMX}$, expressed as:
\begin{equation}
    \mathcal{D}_{1j}^{''}=f_{SMX}(\boldsymbol{\mathfrak{D}})=\frac{exp(\mathcal{D}_{1j}^{'})}{\sum_{\eta=2}^{h}exp(\mathcal{D}_{1\eta}^{'})}.
    \label{smx}
\end{equation}

\textit{LogEig layer}: To facilitate the computation of Eq. (\ref{lem-wrm-2}) in the process of network training, the aforementioned LogEig layer is firstly exploited to transform each $\boldsymbol{v}_j$ into a Euclidean representation residing on the tangent space. Since $\boldsymbol{V}=\{\boldsymbol{\rm{H}}_{h+1},\boldsymbol{\rm{H}}_{h+2},...,\boldsymbol{\rm{H}}_{\mathrm{E}-x+1}\}$, the LogEig function can be described as:
\begin{equation}
\boldsymbol{\bar{\rm{H}}}_t=f_l(\boldsymbol{\rm{H}}_t)=\boldsymbol{U}_t\mathrm{log}(\boldsymbol{\Sigma}_t)\boldsymbol{U}_t^{\rm{T}},
\end{equation}
where $t=h+1\to \mathrm{E}-x+1$, and $\boldsymbol{\rm{H}}_t=\boldsymbol{U}_t\mathrm{log}(\boldsymbol{\Sigma}_t)\boldsymbol{U}_t^{\rm{T}}$ is the eigenvalue decomposition.

\textit{WTS layer}: On the basis of the constructed $SMX$ layer and $LogEig$ layer, the weighted-sum operation involved in Eq. (\ref{lem-wrm-2}) can be realized in this layer via the function $f_{WTS}$. Since the length of $\boldsymbol{V}$, \textit{i.e}, $h-1$, equals to that of $\boldsymbol{\mathfrak{D}}$, we formulate the designed WTS layer as:
\begin{equation}
    \boldsymbol{\Upsilon}=f_{WTS}(\boldsymbol{\tilde{\mathfrak{D}}},\boldsymbol{\bar{V}})=\sum_{j=2}^{h}\mathcal{D}_{1j}^{''}\cdot \boldsymbol{\bar{\rm{H}}}_{j+h-1},
    \label{wts}
\end{equation}
where $\boldsymbol{\tilde{\mathfrak{D}}}=\{\mathcal{D}_{12}^{''},...,\mathcal{D}_{1h}^{''}\}$ and $\boldsymbol{\bar{V}}=\{\boldsymbol{\bar{\rm{H}}}_{h+1},...,\boldsymbol{\bar{\rm{H}}}_{\mathrm{E}-x+1}\}$ are generated by the layers of $SMX$ and $LogEig$, respectively.

\textit{ExpEig layer}: This layer is designed to project the corresponding tangent representation, \textit{i.e.}, $\boldsymbol{\Upsilon}$ output by the WTS layer, back onto the SPD manifold. It is realized by the exponential mapping function $f_{\rm{exp}}$, expressed as: 
\begin{equation}
   \boldsymbol{\acute{\Upsilon}}=f_{\rm{exp}}(\boldsymbol{\Upsilon})=\boldsymbol{\grave{U}}\mathrm{exp}(\boldsymbol{\grave{\Sigma}})\boldsymbol{\grave{U}}^{\rm{T}},
   \label{exp}
\end{equation}
where $\boldsymbol{\Upsilon}=\boldsymbol{\grave{U}}\boldsymbol{\grave{\Sigma}}\boldsymbol{\grave{U}}^{\rm{T}}$ refers to the eigenvalue decomposition, and ${\rm{exp}}(\boldsymbol{\grave{\Sigma}})$ denotes the exponential operation on each diagonal element of $\boldsymbol{\grave{\Sigma}}$. In our design, $\boldsymbol{\acute{\Upsilon}}$ is treated as the learned new manifold-valued representation of the $(\mathrm{E}-x+1)^{\rm{th}}$ hidden layer of SSMAE.

\textbf{Objective Function}: Before introducing the objective function used in this paper, it is necessary to explain what modifications we have made to the original DSPDNet \cite{sdsma}. Firstly, to make the representation learning of the entire network lightweight, the two-stage metric learning framework has been removed from DSPDNet. As studied in \cite{sdsma}, the designed metric learning terms need to step through all the possible computing pairs to construct the intra- and inter-class scatter matrices, which inevitably increases the computational burden. Secondly, the ReCov layer used to regularize the elements of SPD matrix in the $(-\epsilon, 0]$ interval has also been removed. Our main consideration is that the constraints that the ReCov regularization needs to satisfy in maintaining the positive definiteness of the feature matrix \cite{tnnls2} will increase the complexity of network debugging. Finally, also for the purpose of simplifying the network design, we remove the reconstruction error term (RT) associated with each SMAE and instead build only one RT at the input and output of SSMAE. Besides, minimising the signal approximation error will tend to diagonalise the corresponding SPD matrices. As a consequence, the number of variables that contribute to the generation of new feature matrices will be reduced. Therefore, we expect to retain more statistics by reducing the number RTs. As presented in Fig. \ref{fig-1}, the objective function used in this article can be formulated as:
\begin{equation}
    L(\boldsymbol{\Theta}; \boldsymbol{\mathcal{X}})=\sum_{e=1}^{\rm{E}}\sum_{i=1}^{N}\lambda_1\mathcal{L}_e({\boldsymbol{\rm{X}}_i}, l_i) + \sum_{i=1}^N\lambda_2\mathcal{L}(\boldsymbol{R}_i, \boldsymbol{\rm{\hat{H}}}_{\rm{E}}),
    \label{loss}
\end{equation}
where $\boldsymbol{\Theta}$ denotes the network parameters to be learned, $\lambda_1$ and $\lambda_2$ are two trade-off parameters, and $\boldsymbol{\rm{\hat{H}}}_{\rm{E}}$ represents the reconstructed data, computed by: $\boldsymbol{\rm{\hat{H}}}_{\rm{E}}=\boldsymbol{W}_{\mathrm{E}_2}\boldsymbol{\acute{\Upsilon}}\boldsymbol{W}_{\mathrm{E}_2}^{\rm{T}}$.

In Eq. (\ref{loss}), the first term is the cross-entropy loss \cite{sdsma}, which is built on the basis of the LogEig and FC layers. The second term is the reconstruction error term, used to strengthen the similarity between the input and reconstructed samples. In this article, the Euclidean distance (EuD) is adopted for similarity measurement, as shown below:
\begin{equation}
    \mathcal{L}(\boldsymbol{R}_i, \boldsymbol{\rm{\hat{H}}}_{\rm{E}})=||\boldsymbol{R}_i - \boldsymbol{\rm{\hat{H}}}_{\rm{E}}||_{\rm{F}}^2.
\end{equation}
As mentioned in \cite{sdsma}, the fundamental reason for using EuD to replace LEM is the avoiding of eigenvalue operation and matrix inversion. Besides, the study in \cite{tnnls2} also demonstrates the feasibility of this replacement from both theoretical and experimental aspects.

\subsection{Backward Propagation} 
Due to space limitation, this part just gives the gradient computation in the layers of ExpEig, WTS, SMX, SIM, and LEM defined in the aforementioned geometric learning module. For other layers such as BiMap, ReEig, and LogEig, please kindly refer to \cite{sdsma}\cite{spdnet}. Note that the symbol $k$ is used to represent any layer in the designed network for simplicity.

\textit{ExpEig layer}: Similar to the LogEig layer, the ExpEig layer also involves SVD operation. Hence, we learn from the practice of \cite{spdnet} to split its computation into two parts. The first part treats $\boldsymbol{\Upsilon}$ as its input and outputs $\boldsymbol{\grave{U}}$ and $\boldsymbol{\grave{\Sigma}}$, denoted as: $\boldsymbol{\acute{\Upsilon}}=f_{\rm{exp}}^{(k^+)}(\boldsymbol{\Upsilon}_{k-1})=(\boldsymbol{\grave{U}},\boldsymbol{\grave{\Sigma}})$, where $k^+$ signifies the designed transition layer. Since $d\boldsymbol{\Upsilon}=d\boldsymbol{\grave{U}}\boldsymbol{\grave{\Sigma}}\boldsymbol{\grave{U}}^{\rm{T}}+\boldsymbol{\grave{U}}d\boldsymbol{\grave{\Sigma}}\boldsymbol{\grave{U}}^{\rm{T}}+\boldsymbol{\grave{U}}\boldsymbol{\grave{\Sigma}}d\boldsymbol{\grave{U}}^{\rm{T}}$, the partial derivative of $\mathcal{J}^{(k)}$ w.r.t $\boldsymbol{\Upsilon}$ is given below \cite{spdnet}:
\begin{equation}
    \frac{\partial \mathcal{J}^{(k)}}{\partial \boldsymbol{\Upsilon}}=2\boldsymbol{\grave{U}} [\boldsymbol{\Phi}^{\rm{T}}\circ (\boldsymbol{\grave{U}}^{\rm{T}}\boldsymbol{\Omega}_1)]_{sym}\boldsymbol{\grave{U}}^{\rm{T}}+\boldsymbol{\grave{U}}(\boldsymbol{\Omega}_2)_{diag}\boldsymbol{\grave{U}}^{\rm{T}},
    \label{exp-1}
\end{equation}
where $\mathcal{J}^{(k)}=\mathcal{L}_{\rm{E}}\circ f^{(K)}\circ ... \circ f^{(k)}$ represents the loss function of the $k^{\rm{th}}$ layer,  $\boldsymbol{\Omega}_1=\frac{\partial \mathcal{J}^{(k^+)}}{\partial \boldsymbol{\grave{U}}}$, $\boldsymbol{\Omega}_2=\frac{\partial \mathcal{J}^{(k^+)}}{\partial \boldsymbol{\grave{\Sigma}}}$, $\boldsymbol{B}_{sym}=(\boldsymbol{B} + \boldsymbol{B}^{\rm{T}})/2$, and $\boldsymbol{\Phi}_{ij}=1/(\delta_i^2-\delta_j^2)$ if $i\neq j$ and 0 otherwise. Here, $\delta_i$ represents the $i^{\rm{th}}$ ($i=1\to d_{k-1}$)  eigenvalue of $\boldsymbol{\grave{\Sigma}}$.   

The second part receives $\boldsymbol{\grave{U}}$ and $\boldsymbol{\grave{\Sigma}}$ as its inputs and outputs $\boldsymbol{\acute{\Upsilon}}_{k}$ using Eq. (\ref{exp}). Hence, the following variation can be derived:
\begin{equation}
 d\boldsymbol{\acute{\Upsilon}}=2(d\boldsymbol{\grave{U}}\boldsymbol{\grave{\Sigma}}\boldsymbol{\grave{U}}^{\rm{T}})_{sym}+[\boldsymbol{\grave{U}}{\rm{exp}}(\boldsymbol{\grave{\Sigma}})d\boldsymbol{\grave{\Sigma}}\boldsymbol{\grave{U}}^{\rm{T}}]_{sym},
\end{equation}
By referring to the derivation strategy used in the LogEig layer \cite{spdnet}, the partial derivatives of $\mathcal{J}^{(k^+)}$ with respect to $\boldsymbol{\grave{U}}$ and $\boldsymbol{\grave{\Sigma}}$ are given below:
\begin{equation}
    \frac{\partial \mathcal{J}^{(k^+)}}{\partial \boldsymbol{\grave{U}}}=2\boldsymbol{\Delta}\boldsymbol{\grave{U}}{\rm{exp}}(\boldsymbol{\grave{\Sigma}})\;\;\;\;\; \frac{\partial \mathcal{J}^{(k^+)}}{\partial \boldsymbol{\grave{\Sigma}}}={\rm{exp}}(\boldsymbol{\grave{\Sigma}})\boldsymbol{\grave{U}}^{\rm{T}}\boldsymbol{\Delta}\boldsymbol{\grave{U}},
    \label{exp-2}
\end{equation}
where $\boldsymbol{\Delta}=\left(\frac{\partial \mathcal{J}^{(k+1)}}{\partial \boldsymbol{\acute{\Upsilon}}}\right)_{sym}$.

\textit{WTS layer}: Based on Eq. (\ref{wts}), the partial derivatives of $\mathcal{J}^{(k)}$ w.r.t $\mathcal{D}_{1j}^{''}$ and $\boldsymbol{\bar{\rm{H}}}_{j+h-1}$ can be computed by:
\begin{align}
    &\frac{\partial \mathcal{J}^{(k)}}{\partial \boldsymbol{\bar{\rm{H}}}_{j+h-1}} = \mathcal{D}_{1j}^{''} \cdot \frac{\partial \mathcal{J}^{(k+1)}}{\partial \boldsymbol{\Upsilon}} \label{wts-1},\\
    &\frac{\partial \mathcal{J}^{(k)}}{\partial \mathcal{D}_{1j}^{''}}= trace\left(\boldsymbol{\bar{\rm{H}}}_{j+h-1}\frac{\partial \mathcal{J}^{(k+1)}}{\partial \boldsymbol{\Upsilon}}\right).
    \label{wts-2}
\end{align}

\textit{SMX layer}: According to Eq. (\ref{smx}), the partial derivative of $\mathcal{J}^{(k)}$ w.r.t $\mathcal{D}_{1j}^{'}$ is computed as:
\begin{equation}
    \frac{\partial \mathcal{J}^{(k)}}{\partial \mathcal{D}_{1j}^{'}}=\frac{exp(\mathcal{D}_{1j}^{'})\cdot Q}{[\sum_{\eta=2}^h exp(\mathcal{D}_{1\eta}^{'})]^2}\cdot \frac{\partial \mathcal{J}^{(k+1)}}{\partial \mathcal{D}_{1j}^{''}},
    \label{smx-1}
\end{equation}
where $Q = [\sum_{\eta=2}^h exp(\mathcal{D}_{1\eta}^{'}) - exp(\mathcal{D}_{1j}^{'})]$.

\textit{SIM layer}: Based on Eq. (\ref{sim}), the partial derivative of $\mathcal{J}^{(k)}$ w.r.t $\mathcal{D}_{1j}$ can be computed as follows:
\begin{equation}
    \frac{\mathcal{J}^{(k)}}{\mathcal{D}_{1j}}= - \frac{1}{[1+log(1+\mathcal{D}_{1j})]^2}\cdot \frac{1}{(1+\mathcal{D}_{1j})}\cdot \frac{\partial \mathcal{J}^{(k+1)}}{\partial \mathcal{D}_{1j}^{'}}.
    \label{sim-1}
\end{equation}

\textit{LEM layer}: According to Eq. (\ref{lem-layer}), the partial derivatives of $\mathcal{J}^{(k)}$ w.r.t $\boldsymbol{\rm{H}}_1$ and $\boldsymbol{\rm{H}}_j$ can be computed by:
\begin{alignat}{1}
    &\frac{\mathcal{J}^{(k)}}{\boldsymbol{\rm{H}}_1} = 2\boldsymbol{\rm{H}}_1^{-1}[\rm{log}(\boldsymbol{\rm{H}}_1)-\rm{log}(\boldsymbol{\rm{H}}_j)]\cdot \frac{\mathcal{J}^{(k+1)}}{\mathcal{D}_{1j}} \label{lem-1},\\
    &\frac{\mathcal{J}^{(k)}}{\boldsymbol{\rm{H}}_j} = -2\boldsymbol{\rm{H}}_j^{-1}[\rm{log}(\boldsymbol{\rm{H}}_1)-\rm{log}(\boldsymbol{\rm{H}}_j)]\cdot \frac{\mathcal{J}^{(k+1)}}{\mathcal{D}_{1j}}.
    \label{lem-2}
\end{alignat}

With the aid of the aforementioned ingredients, the Riemannian matrix backpropagation for network training can be realized. \textbf{Algorithm \ref{alg1}} summarizes the main implementation details of the proposed SMSA-GLM.

\begin{algorithm}
\caption{SMSA-based Geometric Learning Module}
\textbf{Input:} The number of stacked SMAEs $\rm{E}$, query set $\boldsymbol{Q}=\{\boldsymbol{\rm{H}}_1\}$, key set $\boldsymbol{K}=\{\boldsymbol{\rm{H}}_2,...,\boldsymbol{\rm{H}}_h\}$, value set $\boldsymbol{V}=\{\boldsymbol{\rm{H}}_{h+1},...,\boldsymbol{\rm{H}}_{\mathrm{E}-x+1}\}$, and training epochs $\mathcal{E}$.
\begin{algorithmic}
\renewcommand{\algorithmicrequire}{\textbf{Forward Pass:}}
\REQUIRE
\STATE \textbf{\textit{for}} $\tau \leftarrow 1\;\mathrm{\textbf{to}}\;\mathcal{E}$ \textbf{\textit{do}}:\\
\STATE$\;\;\;\textbf{\textit{for}}$ $j\leftarrow 2\;\mathrm{\textbf{to}}\; h$ \textbf{\textit{do}}:\\
$\;\;\;\;\;\;\;\;\;$\textit{case 'LEM layer'}\\

\STATE 
$\;\;\;\;\;\;\;\;\;$$\mathcal{D}_{1j}=f_{LEM}(\boldsymbol{\rm{H}}_1,\boldsymbol{\rm{H}}_j)=||\rm{log}(\boldsymbol{\rm{H}}_1)-\rm{log}(\boldsymbol{\rm{H}}_j)||_{\rm{F}}^2$.\\
$\;\;\;\;\;\;\;\;\;\,$\textit{case 'SIM layer'}

\STATE $\;\;\;\;\;\;\;\;\;\,$$\mathcal{D}_{1j}^{'} =f_{SIM}(\mathcal{D}_{1j})=\frac{1}{1+log(1+\mathcal{D}_{1j})}$.\\

$\;\;\;\;\;\;\;\;\;\,$\textit{case 'SMX layer'}
\STATE 
$\;\;\;\;\;\;\;\;\;\,$$\mathcal{D}_{1j}^{''}=f_{SMX}(\boldsymbol{\mathfrak{D}})=\frac{exp(\mathcal{D}_{1j}^{'})}{\sum_{\eta=2}^{h}exp(\mathcal{D}_{1\eta}^{'})}$.\\

$\;\;\;\;\;\;\;\;\;\,$\textit{case 'LogEig layer'}
\STATE 
$\;\;\;\;\;\;\;\;\;\,$$\boldsymbol{\bar{\rm{H}}}_{j+h-1}=f_l(\boldsymbol{\rm{H}}_{j+h-1})=\boldsymbol{U}_{j+h-1}\mathrm{log}(\boldsymbol{\Sigma}_{j+h-1})\boldsymbol{U}_{j+h-1}^{\rm{T}}$.\\

$\;\;\;\;\;\;\;\;\;\,$\textit{case 'WTS layer'}
\STATE 
$\;\;\;\;\;\;\;\;\;\,$$\boldsymbol{\Upsilon}=f_{WTS}(\boldsymbol{\tilde{\mathfrak{D}}},\boldsymbol{\bar{V}})=\sum_{j=2}^{h}\mathcal{D}_{1j}^{''}\cdot \boldsymbol{\bar{\rm{H}}}_{j+h-1}$.\\

$\;\;\;\;\;\;\;\;\;\,$\textit{case 'ExpEig layer'}
\STATE 
$\;\;\;\;\;\;\;\;\;\,$$\boldsymbol{\acute{\Upsilon}}=f_{\rm{exp}}(\boldsymbol{\Upsilon})=\boldsymbol{\grave{U}}\mathrm{exp}(\boldsymbol{\grave{\Sigma}})\boldsymbol{\grave{U}}^{\rm{T}}$.\\
\STATE$\;\;\;\textbf{\textit{end}}$\\
\STATE$ \textbf{\textit{end}}$\\

\renewcommand{\algorithmicrequire}{\textbf{Backward Pass:}}
\REQUIRE
\STATE \textbf{\textit{for}} $\tau \leftarrow 1\;\mathrm{\textbf{to}}\;\mathcal{E}$ \textbf{\textit{do}}:\\
\STATE$\;\;\;\textbf{\textit{for}}$ $j\leftarrow 2\;\mathrm{\textbf{to}}\; h$ \textbf{\textit{do}}:\\
$\;\;\;\;\;\;\;\;\;\,$\textit{case 'ExpEig layer'}\\

\STATE 
$\;\;\;\;\;\;\;\;\;\,$Computing $\frac{\partial \mathcal{J}^{(k)}}{\partial \boldsymbol{\Upsilon}}$ using Eq. (\ref{exp-1}) and Eq. (\ref{exp-2}).\\
$\;\;\;\;\;\;\;\;\;$\textit{case 'WTS layer'}

\STATE $\;\;\;\;\;\;\;\;\;$Computing $\frac{\partial \mathcal{J}^{(k)}}{\partial \boldsymbol{\bar{\rm{H}}}_{j+h-1}}$ and $\frac{\partial \mathcal{J}^{(k)}}{\partial \mathcal{D}_{1j}^{''}}$ using Eqs. (\ref{wts-1})-(\ref{wts-2}).\\

$\;\;\;\;\;\;\;\;\;\,$\textit{case 'LogEig layer'}
\STATE 
$\;\;\;\;\;\;\;\;\;\,$Please kindly refer to \cite{spdnet} for detailed information.\\

$\;\;\;\;\;\;\;\;\;\,$\textit{case 'SMX layer'}
\STATE 
$\;\;\;\;\;\;\;\;\;\,$Computing $\frac{\partial \mathcal{J}^{(k)}}{\partial \mathcal{D}_{1j}^{'}}$ using Eq. (\ref{smx-1}).\\

$\;\;\;\;\;\;\;\;\;\,$\textit{case 'SIM layer'}
\STATE 
$\;\;\;\;\;\;\;\;\;\,$Computing $\frac{\mathcal{J}^{(k)}}{\mathcal{D}_{1j}}$ using Eq. (\ref{sim-1}).\\

$\;\;\;\;\;\;\;\;\;\,$\textit{case 'LEM layer'}
\STATE 
$\;\;\;\;\;\;\;\;\;\,$Computing $\frac{\mathcal{J}^{(k)}}{\boldsymbol{\rm{H}}_1}$ and $\frac{\mathcal{J}^{(k)}}{\boldsymbol{\rm{H}}_j}$ using Eqs. (\ref{lem-1})-(\ref{lem-2}).\\
\STATE$\;\;\;\textbf{\textit{end}}$\\
\STATE$ \textbf{\textit{end}}$\\
\end{algorithmic}
\textbf{Output}: The obtained new feature matrix $\boldsymbol{\acute{\Upsilon}}$ of the hidden layer of the ${\rm{E}}^{\rm{th}}$ SMAE.
\label{alg1}
\end{algorithm}

\section{Experiments}
\label{expe}
In this section, we examine the efficacy of the suggested method\footnote{The source code will be available at: https://github.com/GitWR/SMSA-GLM} on two typical visual classification tasks, which are conducted on two different benchmarking datasets. Specifically, we investigate video-based facial emotion recognition on the AFEW dataset \cite{icmi} and skeleton-based hand action recognition on the FPHA dataset \cite{fpha}. 

\subsection{Implementation Details}
To establish the network structure of the backbone and the $e^{\rm{th}}$ SPD manifold autoencoder (SMAE) used in this article, the following operation layers are stacked together, which are $\boldsymbol{\rm{X}}_i\to f_b^{(1)}\to f_{r}^{(2)}\to f_b^{(3)}\to f_{r}^{(4)}\to f_b^{(5)}$ and $f_{b}\;{(\rm{input})}\to f_{r}\to f_{b}\;{(\rm{hidden})}\to f_{b}\; {(\rm{reconstruction})}$, respectively. Wherein, $f_b$ and $f_{r}$ respectively represent the layers of BiMap and ReEig. Note that the aforementioned network structure is exactly the same as that in \cite{sdsma}, with the only difference being that all the ReCov layers are removed. As mentioned earlier, this is done to ease the difficulty of debugging the network. Besides, the hidden layer of each SMAE also connects to a classification module, making up of three functional layers: $f_l\,({\rm{LogEig\; layer}})\to f_{fc}\, ({\rm{FC\;layer}})\to f_s\,({\rm{Corss-Entropy\;loss}})$.  
In this paper, we configure the following two important network parameters, \textit{i.e.}, the learning rate $\xi$ and the activation threshold $\epsilon$, as 0.01 and 1$e$-4 on the AFEW and FPHA datasets. Besides, we make $\xi$ attenuate by a factor of 0.8 every 50 epochs on the AFEW dataset. The trade-off parameters $\lambda_1$ and $\lambda_2$ defined in Eq. (\ref{loss}) are another two pivotal parameters, which will be discussed later.
Considering that the computation of SPD networks can not be accelerated by GPU (a series of eigenvalue operations seems to be the main bottleneck) \cite{spdnetbn}, a PC equipped with an i7-9700 CPU (3.4GHz) and 16GB RAM is exploited to train the network. In addition, the batch size $\mathbbm{B}$ is set to 30 on the AFEW and FPHA datasets. 

\subsection{Comparative Methods and Settings}
In this paper, several representative Riemannian manifold learning-based visual classification methods are selected to better evaluate the effectiveness of the proposed approach, which can be grouped into the following four categories: 1) \textit{general methods for SPD matrix learning}, including Log-Euclidean Metric Learning (LEML) \cite{leml} and SPD Manifold Learning (SPDML) \cite{spdml}; 2) \textit{general methods for linear subspace learning}, including Projection Metric Learning (PML) \cite{pml}, Graph Embedding Projection Metric Learning (GEPML) \cite{gepml}, and Graph Embedding Multi-Kernel Metric Learning (GEMKML) \cite{gemkml}; 3) \textit{multi-order statistical learning methods}, containing Hybrid Euclidean-and-Riemannian Metric Learning (HERML) \cite{herml} and Multiple Riemannian Manifolds Metric Learning (MRMML) \cite{mrmml}; 4) \textit{Riemannian deep learning methods}, containing Grassmannian Neural Network (GrNet) \cite{grnet}, SPD Neural Network (SPDNet) \cite{spdnet}, SPDNet embedded with Riemannian Batch Normalization (SPDNetBN) \cite{spdnetbn}, Lightweight SPD Neural Network (SymNet) \cite{symnet}, Manifold-valued Deep Network (ManifoldNet) \cite{manifoldnet}, and our baseline model, \textit{i.e.}, Deep SPDNet (DSPDNet) \cite{sdsma}. 

In the experiments, we run the official implementations of the aforementioned comparative methods to obtain their classification scores on the AFEW and FPHA datasets. 
Since the input SPD data points of ManifoldNet need to have multiple channels, it can not be applied to skeleton datasets such as FPHA.    
For a fair comparison, the key parameters of each method are determined in accordance with the recommendations given by the original authors, and we report the current best results for those competitors that leveraging cross-validation for parameter selection. To be specific, we select the proper values of $\zeta$ and $\eta$ for LEML in the sets of $[0.1:0.1:1]$ and $[0.1,1,10]$, respectively. In SPDML, GEPML, and GEMKML, the number of intra-class and inter-class nearest neighbors of a given sample used to construct the feature graph are selected by cross-validation. For PML, the value of the trade-off parameter $\alpha$ is tuned using cross-validation. In HERML, the suitable values of $\zeta$ and $\gamma$ are searched in the scopes of $[0.1:0.1:1]$ and $[0.001,0.01,0.1,1,10,100,100]$, respectively. For MRMML, the dimensionality $d_w$ of the learned feature subspace for classification is set to be the same as \cite{mrmml}. 
On the AFEW dataset, the learning rate $\lambda$, activation threshold $\epsilon$, and sizes of the weight matrices of SPDNet and SPDNetBN are configured as $0.01$, $1e-4$, and $\{400\times 200, 200\times 100, 100\times 50\}$, and those of GrNet (excluding $\epsilon$) are set to $0.01$ and $\{400\times 300, 300\times 150\}$. The values of these parameters on the FPHA dataset are determined by cross-validation. 
Besides, the batch size $\mathbbm{B}$ of SPDNet, SPDNetBN, and GrNet is set to 30 on all the involved datasets. 
For SymNet, the settings of the activation threshold $\epsilon$, regularization coefficient $\eta$, and kernel sizes are consistent with \cite{symnet}. For DSPDNet, we run it with default settings on all the used datasets.  

\subsection{Dataset Description and Settings}
\begin{figure}[!t]
 \centering
 \includegraphics[scale=0.50]{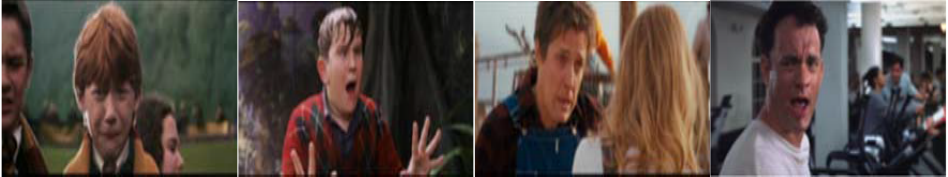}
 \caption{Some facial images of the AFEW dataset}
 \label{fig-afew}
\end{figure}
\begin{figure}[!t]
 \centering
 \includegraphics[scale=0.93]{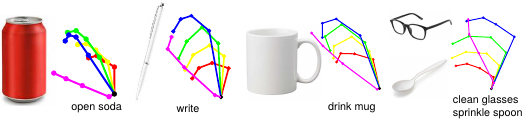}
 \caption{Some skeleton instances of the FPHA dataset}
 \label{fig-fpha}
\end{figure}

\textbf{AFEW dataset:}\;This dataset makes up of 1,345 video sequences about facial expressions collected from movies with close to real-world scenarios. Fig. \ref{fig-afew} shows some facial instances of this dataset. For the evaluation, we follow the data processing guidelines given in \cite{spdnet} to disassemble these training videos into 1,746 small segments for data augmentation, firstly. Then, each frame in the video clip is transformed into a gray-scale image of size $20\times 20$. As a consequence, it is possible to obtain a set of 1,746 SPD matrices with size of $400 \times 400$ for video characterization. Finally, due to the unavailability of publicly accessible groundtruth of the test set, we hereby evaluate the accuracy on the validation set.
On the AFEW dataset, the sizes of the transformation matrices are set to $\{400\times 200,200\times 100\}$ and $\{100\times 50, 50\times 100\}$ for the backbone and the $e^{\rm{th}}$ SMAE, respectively.  

\textbf{FPHA dataset:}\;This dataset serves as a benchmark for estimating hand actions, consisting of 1,175 video sequences of the first person view distributed across 45 distinct categories. A number of hand action examples of this dataset are illustrated in Fig. \ref{fig-fpha}. For a fair comparison, following the established protocol in \cite{fpha}, we first convert each skeleton frame into a feature vector of size $63\times 1$ utilizing the provided 3D coordinates of 21 hand joints. In this context, a total of 1,175 SPD matrices sized $63\times 63$ are computed to represent the action sequences. Subsequently, this dataset is split using a 1:1 ratio, where 600 skeleton-based video clips are allocated for training and the remaining 575 are used as probes. On the FPHA dataset, the sizes of the connection weights w.r.t the backbone are configured as $63\times 53$ and $53\times 43$, and those for the $e^{\rm{th}}$ SMAE are set to $43\times 33$ and $33\times 43$, respectively.

\begin{table}[!t]
\renewcommand\arraystretch{1.1}
   \centering
      \caption{Accuracy Comparison (\%) on the AFEW dataset}
       \vskip -0.05in
       \label{tab-afew-1}
        \begin{tabular}{l|c|c}        
        \hline
           Methods  & Source & AFEW \\ 
          \hline 
           PML \cite{pml} & CVPR'15 & 28.98\\
           LEML \cite{leml} & ICML'15  & 25.13\\
           HERML \cite{herml} & PR'15 & 32.14\\
           MRMML \cite{mrmml} & TBD'22 & 35.71\\
           SPDML-AIM\cite{spdml} & TPAMI'18 & 26.72\\
           SPDML-Stein\cite{spdml} & TPAMI'18 & 24.55\\
           GEPML \cite{gepml} & TCDS'22 & 33.78\\
           GEMKML \cite{gemkml} & TMM'21 & 35.71\\
           GrNet \cite{grnet} & AAAI'18 & 34.23\\
           SPDNet \cite{spdnet} & AAAI'17 & 34.23\\
           SPDNetBN \cite{spdnetbn} & NeurIPS'19 & 36.12\\
           SymNet \cite{symnet} & TNNLS'22 & 32.70\\
           \hline
           DSPDNet-E5 \cite{sdsma} & TCSVT'22 & 37.79 \\
           DSPDNet-E10 \cite{sdsma} & TCSVT'22 & 36.82\\
           \textbf{DSPDNet-SMSA-E5} & & \textbf{38.60 ($\uparrow 0.81$)} \\
           \textbf{DSPDNet-SMSA-E10} & & \textbf{38.65 ($\uparrow 1.83$)} \\
            \hline  
	  \end{tabular}
\end{table}

\begin{table}[!t]
\renewcommand\arraystretch{1.1}
   \centering
      \caption{Accuracy Comparison (\%) on the FPHA dataset}
       \vskip -0.05in
       \label{tab-fpha}
       \resizebox{\linewidth}{!}{
        \begin{tabular}{l|c|ccc|c}
        \hline
           Methods  & Source & Color & Depth & Pose & Acc. \\ 
          \hline 
          JOULE-pose \cite{joule} & CVPR'15 &\ding{55} & \ding{55} & \ding{51} & 74.60 \\
          JOULE-all \cite{joule} & CVPR'15 &\ding{51} & \ding{51} & \ding{51} & 78.78 \\
           Lie Group \cite{lie} & CVPR'14 &\ding{55} & \ding{55} & \ding{51} & 82.69 \\
           Novel View \cite{view}  & CVPR'16 &\ding{55}  & \ding{51}  & \ding{55} & 69.21 \\
           HRNN \cite{hbrnn}  & CVPR'15 &\ding{55} & \ding{55} & \ding{51} & 77.40 \\
           Gram Matrix \cite{gram} & CVPR'16 &\ding{55} & \ding{55} & \ding{51} & 85.39 \\
           Two Stream \cite{ts} & CVPR'16 & \ding{51} & \ding{55} & \ding{55} & 75.30 \\
           TCN \cite{tcn} & CVPRW'17 &\ding{55} & \ding{55} & \ding{51} & 78.57 \\
           TF \cite{tf} & CVPR'17 &\ding{55} & \ding{55} & \ding{51} & 80.69 \\
           LSTM \cite{fpha} & CVPR'18 &\ding{55} & \ding{55} & \ding{51} & 80.14 \\
           ST-GCN \cite{stgcn} & AAAI'18 &\ding{55} & \ding{55} & \ding{51} & 81.30 \\
           TTN \cite{ttn} & CVPR'19 &\ding{55} & \ding{55} & \ding{51} & 83.10 \\
           H+O \cite{ho} & CVPR'19 &\ding{51} & \ding{55} & \ding{55} & 82.43 \\
           LEML \cite{leml} & ICML'15 &\ding{55} & \ding{55} & \ding{51} & 79.48 \\
           SPDML\cite{spdml} & TPAMI'18 &\ding{55} & \ding{55} & \ding{51} & 76.52 \\
           MRMML \cite{mrmml} & TBD'22 &\ding{55} & \ding{55} & \ding{51} &  83.33\\
           GEMKML \cite{gemkml} & TMM'21 &\ding{55} & \ding{55} & \ding{51} & 81.75 \\
           GrNet \cite{grnet} & AAAI'18 &\ding{55} & \ding{55} & \ding{51} & 77.57 \\
           SPDNet \cite{spdnet} & AAAI'17 &\ding{55} & \ding{55} & \ding{51} & 86.26 \\
           SPDNetBN \cite{spdnetbn} & NeurIPS'19 &\ding{55} & \ding{55} & \ding{51} & 86.83 \\
           SymNet \cite{symnet} & TNNLS'22 &\ding{55} & \ding{55} & \ding{51} & 82.96 \\
           \hline
           DSPDNet-E5 \cite{sdsma} & TCSVT'22 &\ding{55} & \ding{55} & \ding{51} & 88.26 \\
           DSPDNet-E10 \cite{sdsma} & TCSVT'22 &\ding{55} & \ding{55} & \ding{51} & 87.97 \\
           \textbf{DSPDNet-SMSA-E5}&  &\ding{55} & \ding{55} & \ding{51} & \textbf{88.83} ($\uparrow 0.57$) \\
           \textbf{DSPDNet-SMSA-E10} &  &\ding{55} & \ding{55} & \ding{51} & \textbf{88.66} ($\uparrow 0.69$) \\
            \hline  
	  \end{tabular}}
\end{table}

\subsection{Results and Discussions}
The classification scores of the different methods achieved on the AFEW dataset are shown in Table \ref{tab-afew-1}. As mentioned in Section \ref{smsa-model} (the part of objective function), the DSPDNet (baseline model) used in this paper can be treated as a lightweight version compared with the original network \cite{sdsma}. Here, 'DSPDNet-E5/E10' signifies that the number E of the stacked SMAEs contained in DSPDNet is 5/10, and DSPDNet-SMSA-E5/E10 embeds the SMSA-based geometric learning model. According to this table, it can be found that the accuracy of DSPDNet-SMSA-E5 is 0.81\% higher than that of the 47-layer DSPDNet-E5. In this article, a deeper model is further explored, \textit{i.e.}, DSPDNet-SMSA-E10, of over 110 layers. As can be seen from Table \ref{tab-afew-1} that this 112-layer DSPDNet-SMSA-E10 achieves a fairly good accuracy on the AFEW dataset (38.65\%, 1.83\% higher than that of the 97-layer DSPDNet-E10). Besides, the classification gap between the deeper network and its corresponding shallower counterpart is shrunk with the aid of the proposed SPD manifold self-attention mechanism (from 0.97\% to 0.05\%). These experimental results demonstrate that the designed SMSA-GLM can effectively mine the complementarity between the geometric features from different network layers, thereby facilitating the generation of more informative deep representations.    

\begin{table}[!t] 
\renewcommand\arraystretch{1.0}
\centering
\caption{Comparison on the AFEW dataset.}
\label{tab-c1}
\begin{tabular}{l|c|c}
\hline
Methods  & \#params & training time (s/epoch) \\
\hline
DSPDNet-E5  & 0.24M & 40.66 \\
DSPDNet-SMSA-E5  & 0.24M & 98.47\\
DSPDNet-E10 & 0.38M & 69.85 \\
DSPDNet-SMSA-E10 & 0.38M & 168.31 \\
\hline
\end{tabular}
\end{table}

In addition to the competitors listed in Table \ref{tab-afew-1}, we also choose several representative action recognition methods for better comparison on the FPHA dataset, which are the jointly learning heterogeneous features (JOULE) \cite{joule}, Lie Group \cite{lie}, Novel View \cite{view}, two stream network (Two Stream) \cite{ts}, hierarchical recurrent neural network (HRNN) \cite{hbrnn}, Gram Matrix \cite{gram}, temporal convolutional network (TCN) \cite{tcn}, transition forests (TF) \cite{tf}, LSTM \cite{fpha}, spatial-temporal graph convolutional network (ST-GCN) \cite{stgcn}, temporal transformer network (TTN) \cite{ttn}, and unified hand and object model (H+O) \cite{ho}. According to the experimental results given in Table \ref{tab-fpha}, we can observe that the classification ability of DSPDNet-SMSA-E5 is also superior to that of DSPDNet-E5. With increasing the network depth, DSPDNet-E10 has performance attenuation compared to DSPDNet-E5. In contrast, the suggested SMSA-GLM is capable of endowing the deeper SPD network with additional capacity of nonlinear learning and discriminative representation, thus being able to mitigate the degradation issue and achieve accuracy gains (DSPDNet-SMSA-E10 scores 0.69\% higher than DSPDNet-E10 and only 0.17\% lower than DSPDNet-SMSA-E5).

\begin{table}[!t] 
\renewcommand\arraystretch{1.0}
\centering
\caption{Comparison on the FPHA dataset.}
\label{tab-c2}
\begin{tabular}{l|c|c}
\hline
Methods  & \#params & training time (s/epoch) \\
\hline
DSPDNet-E5  & 0.26M & 3.81  \\
DSPDNet-SMSA-E5  & 0.26M & 12.52 \\
DSPDNet-E10 & 0.52M & 6.63 \\
DSPDNet-SMSA-E10 & 0.52M & 23.80 \\
\hline
\end{tabular}
\end{table}

To evaluate the computational efficiency of the designed SMSA-GLM, we count the number of parameters (\#params) to learn and the training time of DSPDNet-SMSA-E5/E10 on the AFEW and FPHA datasets in Table \ref{tab-c1} and Table \ref{tab-c2}, respectively. From these two tables, we can note that our SMSA-GLM does not introduce additional parameters to the network learning process. Nevertheless, a series of eigenvalue operations embodied in SMSA-GLM will inevitably increase the computational burden of the Riemannian network used in this article. The experimental results listed in Table \ref{tab-c1} and Table \ref{tab-c2} verify this, \textit{i.e.}, DSPDNet-SMSA-E5/E10 need to spend more training time than the corresponding DSPDNet-E5/E10 on the two used datasets. As one of the possible future works, we plan to explore approximate computation strategies for eigenvalue operation to enable GPU acceleration.  

All in all, the experimental observations mentioned above demonstrate that 
our refurbishment over the original DSPDNet is effective in alleviating the degradation problem of SPD networks. In what follows, some other experiments are carried out to further study the significance of the proposed SMSA-GLM.

\begin{figure*}[!t]
 \centering
 \includegraphics[scale=0.50]{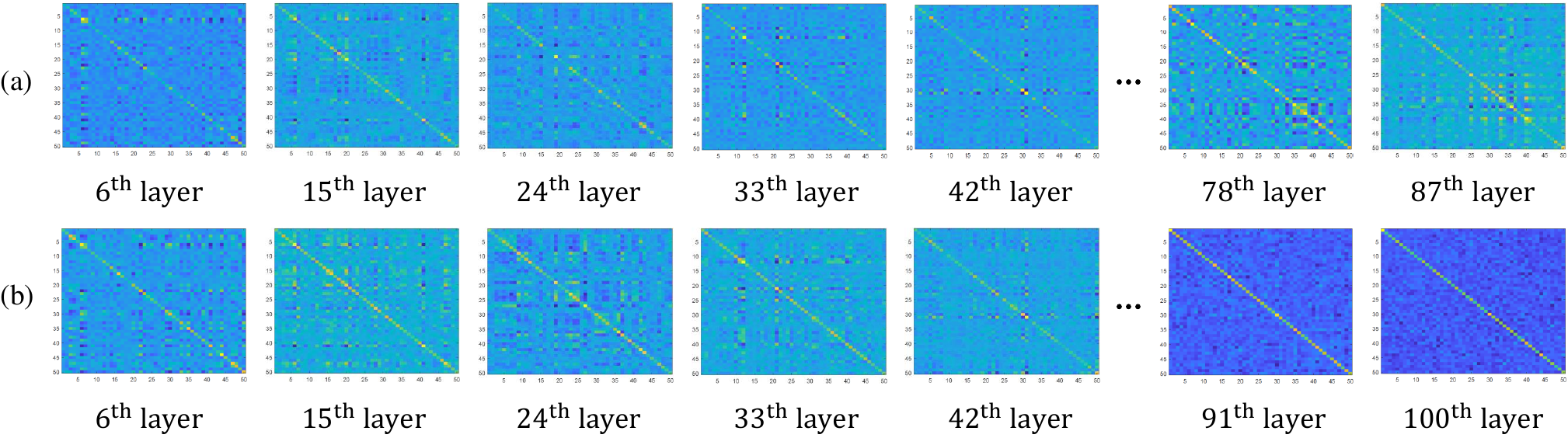}
 \caption{The SPD feature maps generated by different hidden layers of DSPDNet-E10 and DSPDNet-SMSA-E10 on the AFEW dataset are visualized in (a) and (b), respectively. Since the SMSA-GLM is embedded into DSPDNet in this article, the 78$^{\rm{th}}$ and 87$^{\rm{th}}$ layers of (a) become the 91$^{\rm{th}}$ and 100$^{\rm{th}}$ layers shown in (b).}
 \label{fig-vis}
\end{figure*} 

\subsection{Visualization}
To intuitively demonstrate the validity of the proposed SPD manifold self-attention mechanism in facilitating DSPDNet to learn more useful statistical representations, we select the AFEW dataset as an example to visualize the feature maps produced by different hidden layers of DSPDNet-E10 and DSPDNet-SMSA-E10, respectively. According to the results depicted in Fig. \ref{fig-vis}, we can note that these SPD feature maps tend to be diagonalized. 
The fundamental reason can be deduced from a widely acknowledged fact that the optimal solution to the problem of minimising the signal approximation error using a reduced number of basis functions are the eigenvectors that correspond to the largest eigenvalues of the signal covariance matrix.
The problem of reconstructing the SPD matrix (the reconstruction error term is one of the learning objectives of our method) serves as a surrogate for the task of signal approximation. However, by incorporating the designed SMSA-GLM with DSPDNet, it can be found that the concentration of energy on the main diagonal of the feature matrices illustrated in Fig. \ref{fig-vis}(b) generally becomes more pronounced than those shown in Fig. \ref{fig-vis}(a) as the network depth increases. We argue that this is mainly attributed to the enhanced information flow induced by SMSA-GLM, facilitating the reconstruction of the remaining structural information of the input data. In consequence, the pivotal data variations incarnated in the original data points could be captured in the hidden layers, which is beneficial for classification.  

\subsection{Ablation Study for the Proposed SMSA}
\begin{figure}[!t]
 \centering
 \includegraphics[scale=0.58]{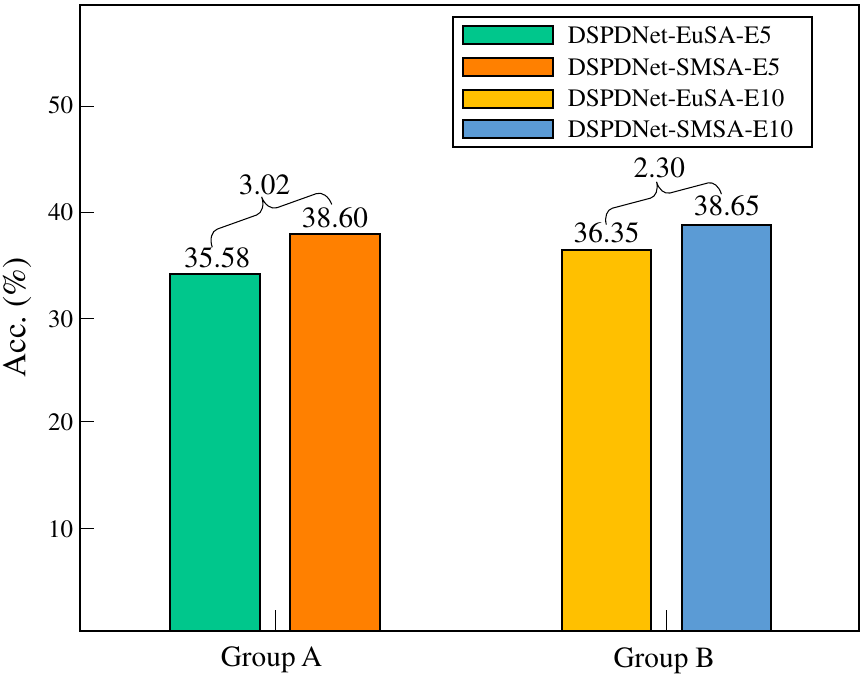}
 \caption{Accuracy comparison (\%) on the AFEW dataset, where 'EuSA' represents the SPD manifold self-attention mechanism implemented by Euclidean computations.}
 \label{fig-eusam}
\end{figure}

Considering that the proposed SPD manifold self-attention mechanism could also be realized using the distance measure and weighted average operation defined in the Euclidean space, we make experiments on the AFEW dataset to test the classification capacity of DSPDNet-EuSMA-E5 and DSPDNet-EuSMA-E10. For simplicity, we abbreviate the Euclidean self-attention mechanism as EuSA. It should be emphasized that the conventional self-attention computation based on ${Softmax}\left(\frac{QK^{\rm{T}}}{\sqrt{d_k}}\right)V$ is not applicable to the SPD data points, because the positive definiteness of the feature matrices will be destroyed. Therefore, for the implementation of EuSA, we first use $\mathcal{D}_{1j}=||\boldsymbol{\rm{H}}_1-\boldsymbol{\rm{H}}_j||_{\rm{F}}^2$ to replace Eq. (\ref{lem-layer}) for similarity measurement. Then, we rewrite Eq. (\ref{wts}) as $\acute{\boldsymbol{\Upsilon}}=\sum_{j=2}^{h}\mathcal{D}_{1j}^{''}\cdot \boldsymbol{\rm{H}}_{j+h-1}$ to obtain the final output feature matrix of SMSA-GLM. Note that the Riemannian geometry of the input data manifold can be preserved under the aforementioned Euclidean computations. From Fig. \ref{fig-eusam}, it is evident that the classification scores of DSPDNet-EuSA-E5 and DSPDNet-EuSA-E10 are respectively 3.02\% and 2.30\% lower than those of DSPDNet-SMSA-E5 and DSPDNet-SMSA-E10 on the AFEW dataset. We argue that the basic reason is the Euclidean distance can not precisely estimate the geodesic similarity between any two SPD elements, making the obtained Riemannian barycenter will deviate from the actual one computed by LEM-based wFM. This certifies that the utilization of Riemannian computations in the proposed SMSA is reasonable and effective.  

\subsection{Convergence Behavior}
\begin{figure}[!t]
 \centering
 \includegraphics[scale=0.57]{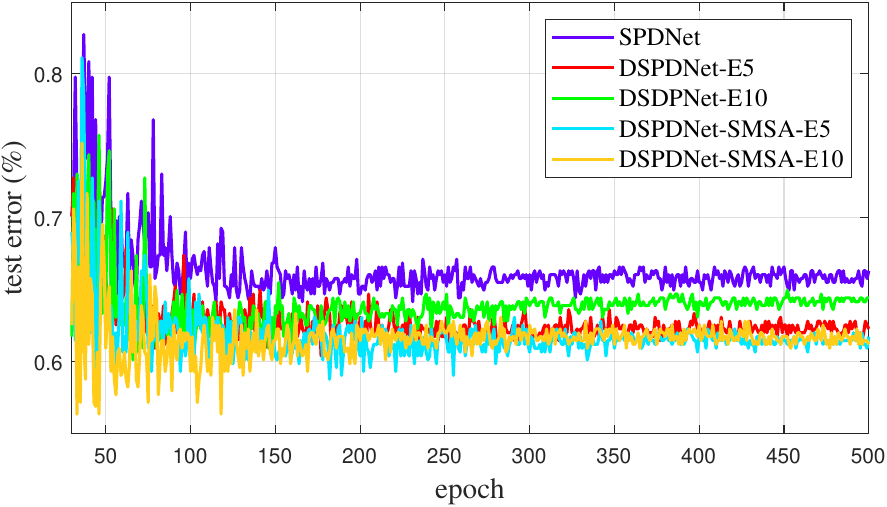}
 \caption{The test error of SPDNet, DSPDNet-E5, DSPDNet-E10, DSPDNet-SMSA-E5, and DSPDNet-SMSA-E10 versus the number of training epochs on the AFEW dataset.}
 \label{fig-covg}
\end{figure}

In this subsection, we select the AFEW dataset as an example to study the convergence behavior of the proposed method. For better comparison, the original SPDNet and our baseline model, \textit{i.e.}, DSPDNet, are chosen as the comparative methods. The classification error of SPDNet, DSPDNet-E5, DSPDNet-E10, DSPDNet-SMSA-E5, and DSPDNet-SMSA-E10 versus the number of training epochs on the AFEW dataset are presented in Fig. \ref{fig-covg}. From this figure, it can be noted that DSPDNet-SMSA-E5 is easy to train, and its convergence speed is still faster than that of SPDNet. When increasing the depth of the network to 112 layers, the convergence performance of DSPDNet-SMSA-E10 shows no sign of fading, compared with DSPDNet-E10. According to Fig. \ref{fig-covg}, we can also find that the number of epochs required for DSPDNet-SMSA-E5 and DSPDNet-SMSA-E10 to reach convergence is within 500. Another important observation from Fig. \ref{fig-covg} is that the degradation phenomenon reflected by the comparison between DSPDNet-E5 and DSPDNet-E10 (after the 200$^{\rm{th}}$ epoch, the green line is distributed above the red line) is effectively alleviated with the aid of the designed SMSA-GLM, \textit{i.e.}, there is no significant positional difference between the cyan and orange lines as the number of epochs increases. In this paper, we set the maximum of training epochs of the designed networks on the AFEW and FPHA datasets to 500 and 1400, respectively.

\subsection{Ablation study for the trade-off parameters $\lambda_1$ and $\lambda_2$ in Eq. (\ref{loss})}
In this subsection, some experiments are carried out using the AFEW and FPHA datasets to explore the influence of the trade-off parameters $\lambda_1$ and $\lambda_2$ on the accuracy of the proposed method, selecting DSPDNet-SMSA-E5 as an illustrative example. In actuality, $\lambda_2$ is employed to strike a balance between the magnitude of the reconstruction error term (RT) and the classification term, thereby facilitating the training of effective classifiers that leads to improved classification results. Consequently, in this paper, the value of $\lambda_1$ is fixed to 1, and $\lambda_2$ is endowed with a comparatively small value for the sake of fine-tuning the performance of our network. To be specific, we respectively configure the candidate sets of $\lambda_2$ on the FPHA and AFEW datasets as $\rm{\{1e-3,1e-2,1e-1,1.0\}}$ and $\rm{\{1e-4,1e-3,1e-2,1e-1\}}$ to study its regulation mechanism on the model capacity. According to Fig. \ref{fig-lambda}, we can get two interesting observations presented below. The first is that DSPDNet-SMSA-E5 shows a steady trend of first increasing and then decreasing in terms of classification score with the change of $\lambda_2$ on the two used datasets. This is mainly attributed to the fact that a larger value of $\lambda_2$ renders the entire network prone to deep reconstruction learning, which is not conductive to producing discriminative feature representations for classification. In contrast, a smaller value of $\lambda_2$ will weaken the supervisory signal provided by RT, making the proposed learning system unable to effectively resist the information degradation issue. 
Secondly, Fig. \ref{fig-lambda} demonstrates the insensitivity of the proposed model towards this trade-off parameter, thereby corroborating our claim that the incorporation of RT and classification term is beneficial to facilitate our model in generating more useful statistical features for decision making. 

When it comes to parameter selection, our principle is to guarantee that the RT is at least ten times smaller in magnitude compared to the classification term. By adhering to this criterion, the softmax classifier can effectively leverage the gradient information of RT to infer a more reasonable classification hypersphere for distinct categories. In this article, we set the proper value of $\lambda_2$ in DSPDNet-SMSA-E5 and DSPDNet-SMSA-E10 to $\{\rm{1e-1},\rm{1e-3}\}$ and $\{\rm{1e-3},\rm{1e-2}\}$ on the FPHA and AFEW datasets, respectively.  
Given a new dataset, the aforementioned selection rule allows the readers to promptly ascertain the initial value range of $\lambda_2$ based on the values of cross-entropy loss.

\begin{figure}[!t]
\centering
\includegraphics[scale=0.48]{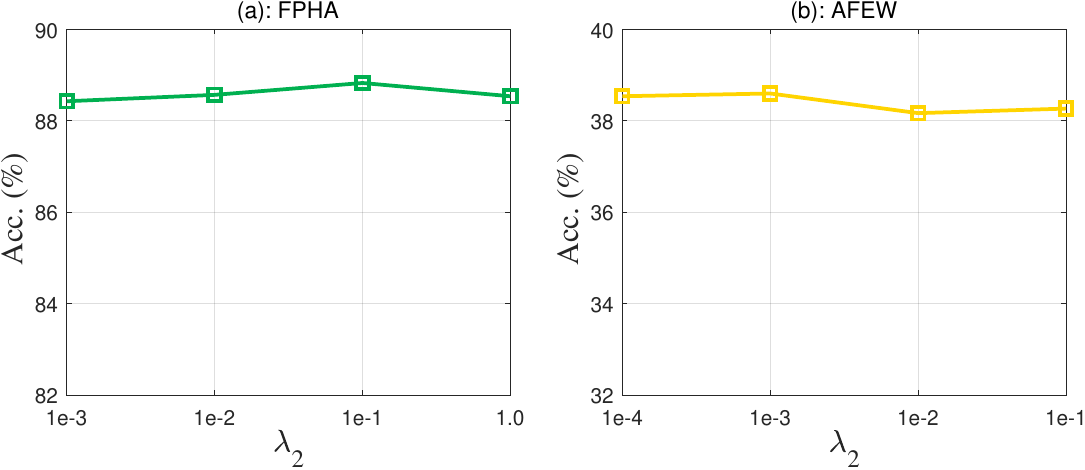}
\caption{Study the impact of the trade-off parameter $\lambda_2$ on the classification accuracy of DSPDNet-SMSA-E5 on the FPHA and AFEW datasets.}
\label{fig-lambda}
\end{figure}

\section{Application to Skeleton-based Human Action Recognition with Unmanned Aerial Vehicles}

Owing to the swift movement and the continuously shifting orientations and heights of unmanned aerial vehicles (UAVs) in flight, the recorded video sequences by UAVs display significant appearance variations in the aspects of clarity, lighting conditions, perspective, and background details. These factors pose considerable challenges for computer vision tasks involving UAVs, such as person re-identification and pose estimation. In this section, we carry out experiments on the large-scale UAV-Human dataset \cite{uav} to further assess the efficacy of the proposed SPD manifold self-attention mechanism in ameliorating the learning capacity of the baseline model.

\begin{figure}[!t]
\centering
\includegraphics[scale=0.48]{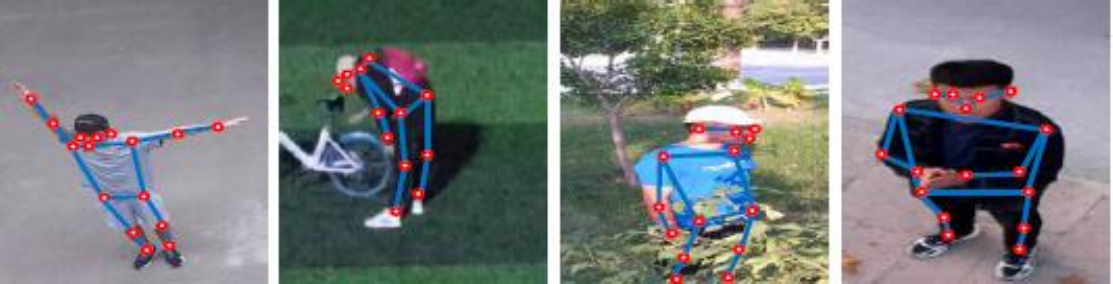}
\caption{Some examples of the UAV-Human dataset.}
\label{uav-fig}
\end{figure}

This dataset comprises a total of 67,428 annotated video sequences for human behavior understanding. Among these, 22,476 videos, spanning 155 action categories, are specifically allocated for the task of human pose estimation.
For a fair comparison, the protocol introduced in \cite{hrgeml} is adopted for data processing. To be specific, we first transform each action frame into a 51-dimensional feature vector, as there are 17 body joints with 3D coordinates labeled in each instance (presented in Fig. \ref{uav-fig}). Moreover, a subset of 305 action frames is selected for computation from each video clip, resulting in that a feature matrix of size $51 \times 305$ can be obtained for video representation. 
However, there are several actions that are completed by two persons in this dataset. Considering that the proportion of this part is relatively small, the PCA technique is then exploited for the projection of such feature vectors (their dimension is 102, \textit{i.e.}, $17\times 3\times 2$) into 51-dimensional ones. Through experiments, it can be found that this dimensionality reduction process can retain 99\% energy of the original data.
With the aforementioned preparations, the original skeleton dataset is now represented by 22,476 SPD matrices of size $51\times 51$ by computing Eq. (\ref{e1}). 
Finally, the seventy-thirty-ratio (STR) criterion is used to create the training and test sets from the randomly selected 16,723 SPD matrices.

It can be learned from the previous analysis that the 112-layer DSPDNet-SMSA (DSPDNet-SMSA-E10) requires a long training time, so this section just takes DSPDNet-SMSA-E5 as an example to carry out comparative experiments.
Here, the sizes of the connection weights of the backbone and the $e^{\rm{th}}$ SMAE are configured as $\{51\times 43,43\times 37\}$ and $\{37\times 31, 31\times 37\}$ on the UAV-Human dataset, respectively.  
Besides, the learning rate $\xi$, batch size $\rm{\mathbbm{B}}$, rectification threshold $\epsilon$, and trade-off parameter $\lambda_2$ are respectively set to 0.01 (attenuate by a factor of 0.8 every 50 epochs), 30, 1$\rm{e}$-5, and $1$\rm{e}$-4$. The recognition scores of DSPDNet-SMSA-E5, DSPDNet-E5, and other competitors are given in Table \ref{tab-uav}. 
Note that we execute the publicly available source codes of these competitors on this dataset, making adjustments to the parameters for achieving optimal accuracy.

\begin{figure}[!t]
\centering
\includegraphics[scale=0.57]{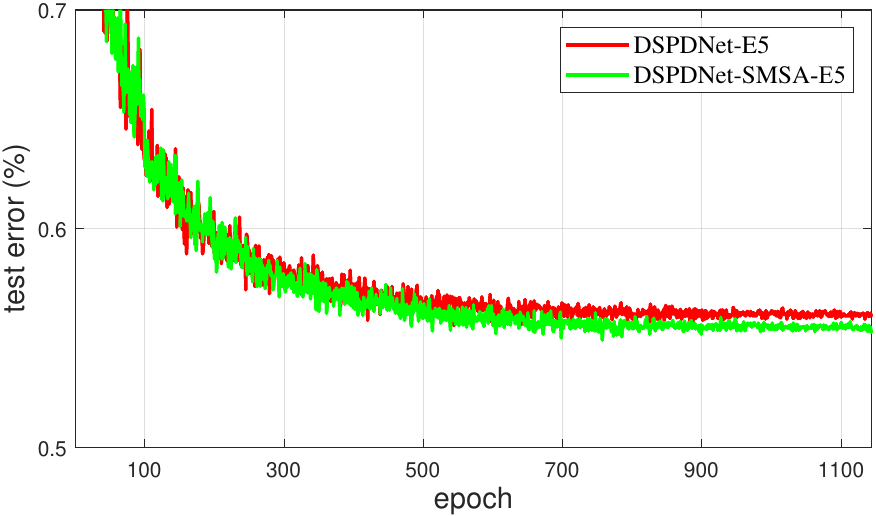}
\caption{Convergence behavior on the UAV-Human dataset.}
\label{uav-covg}
\end{figure}

From Table \ref{tab-uav}, it is evident that DSPDNet-E5 embedded with the designed SMSA-based geometric learning module exhibits the best classification ability. Moreover, compared to the baseline model, DSPDNet-SMSA-E5 achieves a classification score of 0.75\% higher. This further certifies the effectiveness of the proposed SPD manifold self-attention mechanism in harnessing the statistical correlation among geometric features at different layers to facilitate learning a more discriminative manifold-to-manifold embedding mapping. 
Nevertheless, the average training time of DSPDNet-SMSA-E5 lasts for one epoch is 214.02s, which is 138.26s longer than that of DSPDNet-E5. 
In addition, we plot the test error of DSPDNet-SMSA-E5 and DSPDNet-E5 as a function of training epochs on the UAV-Human dataset in Fig. \ref{uav-covg}. It can be observed that the convergence performance of DSPDNet-E5 does not attenuate under the embedding of SMSA-GLM. On the contrary, it is slightly better than the original (the green line is generally below the red line and its convergence speed is relatively faster after the 300$^{\rm{th}}$ epoch). In this part, we set the maximum of training epochs to 1150 accordingly. 
  
\begin{table}[!t]
\renewcommand\arraystretch{1.1}
   \centering
      \caption{Accuracy Comparison (\%) on the UAV-Human dataset}
       \label{tab-uav}
        \begin{tabular}{l|c|c|c}       
        \hline
           Methods  & Source & Pose & Acc. \\ 
          \hline 
          PML \cite{pml} & CVPR'15 & \ding{51} & 10.66 \\
          LEML \cite{leml} & ICML'15 & \ding{51} & 21.83 \\
          HERML \cite{herml} & PR'15 & \ding{51} & 34.18 \\
          HRGEML \cite{hrgeml}  & TBD'23 & \ding{51} & 36.10 \\
          SPDML-AIM \cite{spdml}  & TPAMI'18 & \ding{51} & 22.69 \\
          GEMKML \cite{gemkml} & TMM'21 &\ding{51} & 34.67 \\
          GrNet \cite{grnet} & AAAI'18 & \ding{51} & 35.23 \\
          SPDNet \cite{spdnet} & AAAI'17 & \ding{51} & 42.31 \\
          SPDNetBN \cite{spdnetbn} & NeurIPS'19 & \ding{51} & 43.28 \\
          SymNet \cite{symnet} & TNNLS'22 & \ding{51} & 35.89 \\
           \hline
           DSPDNet-E5 \cite{sdsma} & TCSVT'22 & \ding{51} & 43.98 \\
           \textbf{DSPDNet-SMSA-E5}&  & \ding{51} & \textbf{44.73} ($\uparrow $0.75) \\
            \hline  
	  \end{tabular}
\end{table}

\section{Conclusion}
The problem of structural information degradation in the process of multi-stage feature transformation mapping is treated as a primary factor hindering the SPD networks from going deeper. Although a recent work (DSPDNet \cite{sdsma}) has effectively alleviated this issue by constructing a stacked Riemannian autoencoder network with an identity mapping function, a series of upsampling (output rank-deficient matrices) and nonlinear activation operations w.r.t the SPD matrix inevitably exacerbate the distortion of raw statistics. As a countermeasure, an SPD manifold self-attention mechanism (SMSA) is proposed in this paper to provide a feasible path for capturing manifold-valued long-range relationships. Based on SMSA, we design a geometric learning module for the baseline network (DSPDNet) to intensify its representational capacity by mining the statistical complementarity between different depth features. The experimental results achieved on three benchmarking datasets demonstrate that our proposal is a viable candidate in helping addressing the degradation issue while simultaneously enhancing the classification ability of the original model. In the future, we will focus on the exploration of approximate computational strategies for the eigenvalue problem, so as to improve the training efficiency of SPD networks while maintaining a competitive accuracy. Furthermore, we plan to generalize the paradigm of Euclidean transformers to the context of Riemannian manifolds for the implementation of a coarse-to-fine spatiotemporal modeling and representation.  


\section*{Acknowledgment}

This work was supported by the National Natural Science Foundation of China (62306127, 62020106012, 61672265, 62202205), the Natural Science Foundation of Jiangsu Province (BK20231040), the 111 Project of Ministry of Education of China (B12018), the Fundamental Research Funds for the Central Universities (JUSRP123030), and the UK EPSRC EP/N007743/1, MURI/EPSRC/DSTL EP/R018456/1 grants.

\ifCLASSOPTIONcaptionsoff
  \newpage
\fi



%

\bibliographystyle{IEEEtran}
\bibliography{sample-base}

\end{document}